\documentclass{ieeeaccess}
\usepackage{cite}
\usepackage{amsmath,amssymb,amsfonts}
\usepackage{algorithmic}
\usepackage{graphicx}
\usepackage{textcomp}
\usepackage{xcolor}

\newcommand{\green}[1]{\hlc[green]{#1}}
\newcommand{\yellow}[1]{\hlc[yellow]{#1}}

\renewcommand{\green}[1]{#1}
\renewcommand{\yellow}[1]{#1}

\usepackage{verbatim} 
\usepackage{fancyvrb} 
\usepackage{csquotes}
\usepackage{soul}     
\usepackage[]{cite} 
\usepackage{url} 

\usepackage{amsmath} 
\usepackage{amssymb} 
\usepackage{bm}      
\usepackage[makeroom]{cancel} 

\usepackage[]{graphicx}
\usepackage{graphbox} 
\graphicspath{{images/}{imgs/}} 
\usepackage{subcaption}
\usepackage[font={small}]{caption}
 \usepackage[percent]{overpic}

\usepackage{tabularx}
\usepackage{multirow}
\usepackage{adjustbox} 

\usepackage{arydshln}  
\usepackage{makecell}  
\setcellgapes{4pt}     

\usepackage{rotating}
\usepackage{mathtools}
\usepackage{booktabs}

\def\BibTeX{{\rm B\kern-.05em{\sc i\kern-.025em b}\kern-.08em
    T\kern-.1667em\lower.7ex\hbox{E}\kern-.125emX}}
\begin{document}
\history{}
\doi{}

\title{Anomaly Detection \yellow{in Medical Imaging} with Deep Perceptual Autoencoders}

\author{\uppercase{Nina Shvetsova}\authorrefmark{1,5},
        \uppercase{Bart Bakker}\authorrefmark{2},
        \uppercase{Irina Fedulova}\authorrefmark{1},
        \uppercase{Heinrich Schulz}\authorrefmark{3},
        \uppercase{and~Dmitry V. Dylov}\authorrefmark{4} ~\IEEEmembership{Member,~IEEE}
}
\address[1]{Philips Research, Moscow, Russia}
\address[2]{Philips Research, Eindhoven, Netherlands}
\address[3]{Philips Research, Hamburg, Germany}
\address[4]{Skolkovo Institute of Science and Technology, Moscow, Russia}
\address[5]{Goethe University Frankfurt, Frankfurt, Germany}

\markboth
{Shvetsova \headeretal: Anomaly Detection in Medical Imaging with Deep Perceptual Autoencoders}
{Shvetsova \headeretal: Anomaly Detection in Medical Imaging with Deep Perceptual Autoencoders}

\corresp{Corresponding author: Nina Shvetsova (e-mail: shvetsov@uni-frankfurt.de).}

\begin{abstract}

Anomaly detection is the problem of recognizing abnormal inputs based on the seen examples of normal data. Despite recent advances of deep learning in recognizing image anomalies, these methods still prove incapable of handling complex images, such as those encountered in the medical domain. Barely visible abnormalities in chest X-rays or metastases in lymph nodes on the scans of the pathology slides resemble normal images and are very difficult to detect. To address this problem, we introduce a new powerful method of image anomaly detection. It relies on the classical autoencoder approach with a re-designed training pipeline to handle high-resolution, complex images, and a robust way of computing an image abnormality score. We revisit the very problem statement of fully unsupervised anomaly detection, where no abnormal examples are provided during the model setup. We propose to relax this unrealistic assumption by using a very small number of anomalies of confined variability merely to initiate the search of hyperparameters of the model.
We evaluate our solution on two medical datasets containing radiology and digital pathology images, where the state-of-the-art anomaly detection models, originally devised for natural image benchmarks, fail to perform sufficiently well. 
The proposed approach suggests a new baseline for anomaly detection in medical image analysis tasks\footnote{\green{The source code is available at}  \url{https://github.com/ninatu/anomaly_detection/}}.

\end{abstract}


\titlepgskip=-15pt

\maketitle

\section{Introduction}

\begin{figure*}[tbp]
    \centering
    \begin{subfigure}[]{0.43\textwidth}
        \centering
        \captionsetup{justification=centering, margin=0cm}
        \begin{subfigure}[]{0.125\textwidth}
            \centering
            \includegraphics[width=1\textwidth]{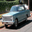}
        \end{subfigure}%
        \begin{subfigure}[]{0.125\textwidth}
            \centering
            \includegraphics[width=1\textwidth]{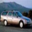}
        \end{subfigure}%
        \begin{subfigure}[]{0.125\textwidth}
            \centering
            \includegraphics[width=1\textwidth]{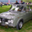}
        \end{subfigure}%
        \begin{subfigure}[]{0.125\textwidth}
            \centering
            \includegraphics[width=1\textwidth]{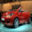}
        \end{subfigure}%
        \begin{subfigure}[]{0.125\textwidth}
            \centering
            \includegraphics[width=1\textwidth]{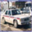}
        \end{subfigure}%
        \begin{subfigure}[]{0.125\textwidth}
            \centering
            \includegraphics[width=1\textwidth]{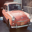}
        \end{subfigure}%
        \begin{subfigure}[]{0.125\textwidth}
            \centering
            \includegraphics[width=1\textwidth]{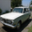}
        \end{subfigure}%
        \begin{subfigure}[]{0.125\textwidth}
            \centering
            \includegraphics[width=1\textwidth]{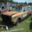}
        \end{subfigure}
    \end{subfigure}%
    \hspace{0.05\textwidth}
    \begin{subfigure}[]{0.43\textwidth}
        \centering
        \captionsetup{justification=centering, margin=0cm}
        \begin{subfigure}[]{0.125\textwidth}
            \centering
            \includegraphics[width=1\textwidth]{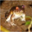}
        \end{subfigure}%
        \begin{subfigure}[]{0.125\textwidth}
            \centering
            \includegraphics[width=1\textwidth]{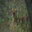}
        \end{subfigure}%
        \begin{subfigure}[]{0.125\textwidth}
            \centering
            \includegraphics[width=1\textwidth]{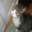}
        \end{subfigure}%
        \begin{subfigure}[]{0.125\textwidth}
            \centering
            \includegraphics[width=1\textwidth]{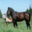}
        \end{subfigure}%
        \begin{subfigure}[]{0.125\textwidth}
            \centering
            \includegraphics[width=1\textwidth]{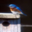}
        \end{subfigure}%
        \begin{subfigure}[]{0.125\textwidth}
            \centering
            \includegraphics[width=1\textwidth]{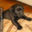}
        \end{subfigure}%
        \begin{subfigure}[]{0.125\textwidth}
            \centering
            \includegraphics[width=1\textwidth]{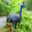}
        \end{subfigure}%
        \begin{subfigure}[]{0.125\textwidth}
            \centering
            \includegraphics[width=1\textwidth]{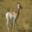}
        \end{subfigure}%
    \end{subfigure}
    \begin{subfigure}[]{0.43\textwidth}
        \centering
        \captionsetup{justification=centering, margin=0cm}
        \begin{subfigure}[]{0.125\textwidth}
            \centering
            \includegraphics[width=1\textwidth]{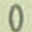}
        \end{subfigure}%
        \begin{subfigure}[]{0.125\textwidth}
            \centering
            \includegraphics[width=1\textwidth]{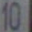}
        \end{subfigure}%
        \begin{subfigure}[]{0.125\textwidth}
            \centering
            \includegraphics[width=1\textwidth]{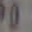}
        \end{subfigure}%
        \begin{subfigure}[]{0.125\textwidth}
            \centering
            \includegraphics[width=1\textwidth]{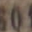}
        \end{subfigure}%
        \begin{subfigure}[]{0.125\textwidth}
            \centering
            \includegraphics[width=1\textwidth]{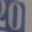}
        \end{subfigure}%
        \begin{subfigure}[]{0.125\textwidth}
            \centering
            \includegraphics[width=1\textwidth]{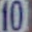}
        \end{subfigure}%
        \begin{subfigure}[]{0.125\textwidth}
            \centering
            \includegraphics[width=1\textwidth]{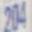}
        \end{subfigure}%
        \begin{subfigure}[]{0.125\textwidth}
            \centering
            \includegraphics[width=1\textwidth]{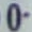}
        \end{subfigure}
    \end{subfigure}%
    \hspace{0.05\textwidth}
    \begin{subfigure}[]{0.43\textwidth}
        \centering
        \captionsetup{justification=centering, margin=0cm}
        \begin{subfigure}[]{0.125\textwidth}
            \centering
            \includegraphics[width=1\textwidth]{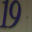}
        \end{subfigure}%
        \begin{subfigure}[]{0.125\textwidth}
            \centering
            \includegraphics[width=1\textwidth]{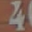}
        \end{subfigure}%
        \begin{subfigure}[]{0.125\textwidth}
            \centering
            \includegraphics[width=1\textwidth]{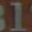}
        \end{subfigure}%
        \begin{subfigure}[]{0.125\textwidth}
            \centering
            \includegraphics[width=1\textwidth]{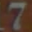}
        \end{subfigure}%
        \begin{subfigure}[]{0.125\textwidth}
            \centering
            \includegraphics[width=1\textwidth]{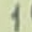}
        \end{subfigure}%
        \begin{subfigure}[]{0.125\textwidth}
            \centering
            \includegraphics[width=1\textwidth]{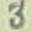}
        \end{subfigure}
        \begin{subfigure}[]{0.125\textwidth}
            \centering
            \includegraphics[width=1\textwidth]{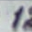}
        \end{subfigure}
        \begin{subfigure}[]{0.125\textwidth}
            \centering
            \includegraphics[width=1\textwidth]{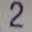}
        \end{subfigure}%
    \end{subfigure}
    \begin{subfigure}[]{0.43\textwidth}
        \centering
        \captionsetup{justification=centering, margin=0cm}
        \begin{subfigure}[]{0.20\textwidth}
            \centering
            \includegraphics[width=1\textwidth]{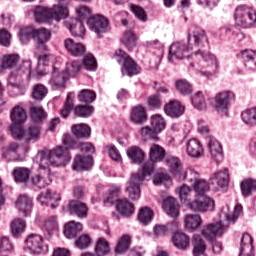}
        \end{subfigure}%
        \begin{subfigure}[]{0.20\textwidth}
            \centering
            \includegraphics[width=1\textwidth]{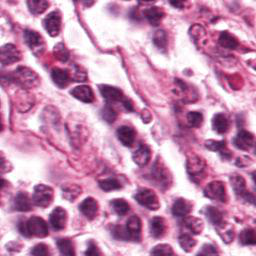}
        \end{subfigure}%
        \begin{subfigure}[]{0.20\textwidth}
            \centering
            \includegraphics[width=1\textwidth]{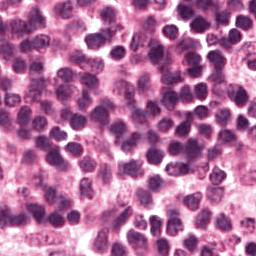}
        \end{subfigure}%
        \begin{subfigure}[]{0.20\textwidth}
            \centering
            \includegraphics[width=1\textwidth]{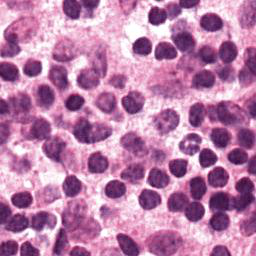}
        \end{subfigure}%
        \begin{subfigure}[]{0.20\textwidth}
            \centering
            \includegraphics[width=1\textwidth]{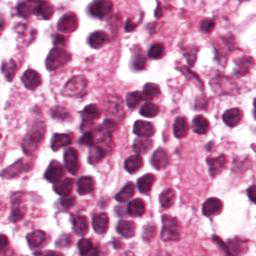}
        \end{subfigure}
    \end{subfigure}%
    \hspace{0.05\textwidth}
    \begin{subfigure}[]{0.43\textwidth}
        \centering
        \captionsetup{justification=centering, margin=0cm}
        \begin{subfigure}[]{0.20\textwidth}
            \centering
            \includegraphics[width=1\textwidth]{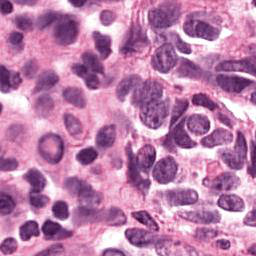}
        \end{subfigure}%
        \begin{subfigure}[]{0.20\textwidth}
            \centering
            \includegraphics[width=1\textwidth]{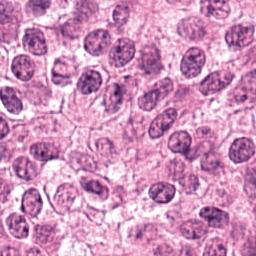}
        \end{subfigure}%
        \begin{subfigure}[]{0.20\textwidth}
            \centering
            \includegraphics[width=1\textwidth]{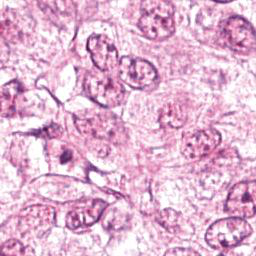}
        \end{subfigure}%
        \begin{subfigure}[]{0.20\textwidth}
            \centering
            \includegraphics[width=1\textwidth]{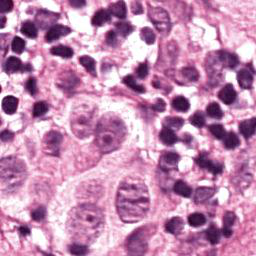}
        \end{subfigure}%
        \begin{subfigure}[]{0.20\textwidth}
            \centering
            \includegraphics[width=1\textwidth]{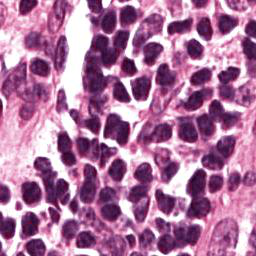}
        \end{subfigure}%
    \end{subfigure}
    \begin{subfigure}[]{0.43\textwidth}
        \centering
        \captionsetup{justification=centering, margin=0cm}
        \begin{subfigure}[]{0.20\textwidth}
            \centering
            \includegraphics[width=1\textwidth]{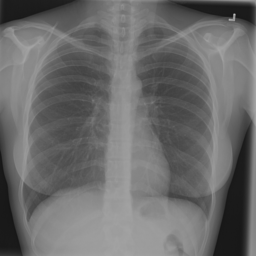}
        \end{subfigure}%
        \begin{subfigure}[]{0.20\textwidth}
            \centering
            \includegraphics[width=1\textwidth]{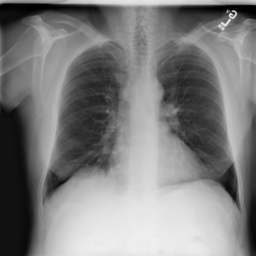}
        \end{subfigure}%
        \begin{subfigure}[]{0.20\textwidth}
            \centering
            \includegraphics[width=1\textwidth]{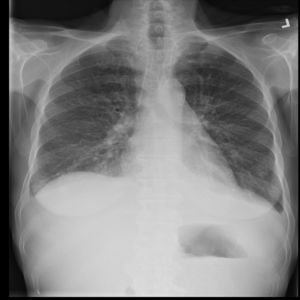}
        \end{subfigure}%
        \begin{subfigure}[]{0.20\textwidth}
            \centering
            \includegraphics[width=1\textwidth]{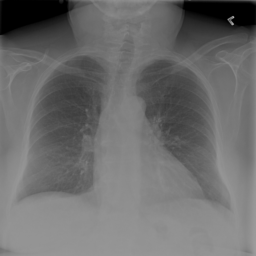}
        \end{subfigure}%
        \begin{subfigure}[]{0.20\textwidth}
            \centering
            \includegraphics[width=1\textwidth]{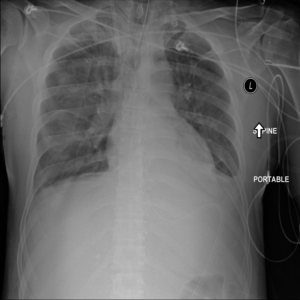}
        \end{subfigure}
    \caption*{\small Normal}
    \end{subfigure}%
    \hspace{0.05\textwidth}
    \begin{subfigure}[]{0.43\textwidth}
        \centering
        \captionsetup{justification=centering, margin=0cm}
        \begin{subfigure}[]{0.20\textwidth}
            \centering
            \includegraphics[width=1\textwidth]{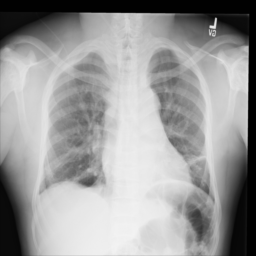}
        \end{subfigure}%
        \begin{subfigure}[]{0.20\textwidth}
            \centering
            \includegraphics[width=1\textwidth]{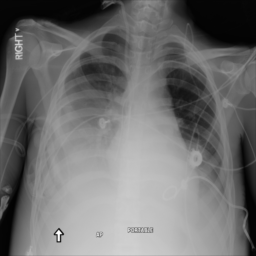}
        \end{subfigure}%
        \begin{subfigure}[]{0.20\textwidth}
            \centering
            \includegraphics[width=1\textwidth]{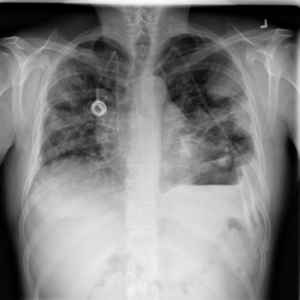}
        \end{subfigure}%
        \begin{subfigure}[]{0.20\textwidth}
            \centering
            \includegraphics[width=1\textwidth]{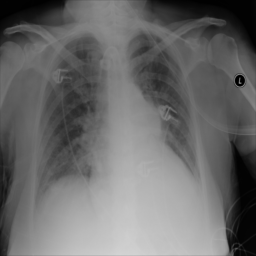}
        \end{subfigure}%
        \begin{subfigure}[]{0.20\textwidth}
            \centering
            \includegraphics[width=1\textwidth]{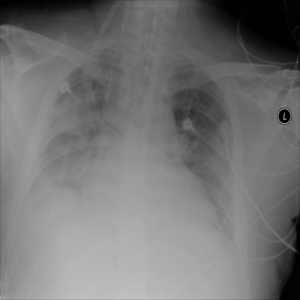}
        \end{subfigure}%
    \caption*{\small Abnormal}
    \end{subfigure}
    \caption{Examples of normal vs. abnormal images of considered datasets. Natural images: (first row) cars vs other classes of CIFAR10 dataset~\cite{krizhevsky2009learning}, (second row) digits ``0'' vs digits ``1''~--~``9'' of SVHN dataset~\cite{netzer2011reading}. Medical images: (third row) healthy tissue vs. tissue with metastases in H\&E-stained lymph nodes images from Camelyon16 challenge~\cite{bejnordi2017diagnostic}, (fourth row) normal chest X-rays vs. chest X-rays with abnormal findings from NIH dataset~\cite{wang2017chestx}. }
    \label{fig:datasets}
\end{figure*}

\IEEEPARstart{A}{nomaly} detection is a crucial task in the deployment of machine learning models, where knowing the ``normal'' data samples should help spot the ``abnormal'' ones~\cite{chandola2009anomaly,chalapathy2019deep}. 
If an input deviates from the training data substantially \yellow{(\textit{e.g.}, the input belongs to a class not represented in the training data)}, it is usually impossible to predict how the model will behave~\cite{nguyen2015deep,amodei2016concrete}. This trait is especially important in high-consequence applications, such as medical decision support systems, where it is especially vital to know how to recognize the anomalous data.
Identification of rare occurrences is another important application where anomaly detection is useful. 
For example, in pathology, where labeling diverse microscopy datasets is both time-consuming and expensive, the rare types of cells and tissues require specialized expertise from the annotator~\cite{Kothari13,Chowdhury17}. 
\green{Forthright anomaly classification and segmentation algorithms are typically prone to mistakes either because of the lack of sufficient annotation (thousands of labeled examples needed for supervised models) or because of the lack of representative data altogether (\textit{e.g.}, the case of some rare pathologies).
Moreover, these algorithms are affected by the need to deal with very unbalanced and \textit{apriori} noisy data, frequently leading to inaccurate results (\textit{e.g.}, the findings on chest x-rays can be so subtle that they can lead to disagreement in the interpretation}~\cite{olatunji2019caveats,ccalli2021deep}).
Because the normal cases greatly prevail over the abnormal ones, the anomaly detection could alleviate the annotation burden by automatically pointing to the rare samples.

In recent years, deep learning techniques achieved important advances in image anomaly detection~\cite{zhai2016deep,gong2019memorizing,schlegl2017unsupervised,perera2019ocgan,golan2018deep,tuluptceva2019perceptual,ouardini2019towards,ruff2018deep,baur2021autoencoders}. 
However, these efforts were primarily focused on artificial problems with distinct anomalies in natural images (e.g., outliers in images of ``cars'' in the CIFAR10 dataset~\cite{krizhevsky2009learning}, see Figure~\ref{fig:datasets}). The medical anomalies, however, differ from those in the natural images~\cite{schlegl2017unsupervised,ouardini2019towards,tang2019deep}. 
\yellow{Contrary to the natural images, the anomalies in the medical domain tend to strongly resemble the normal data.}
For example, detection of obscure neoplasms in chest X-rays~\cite{wang2017chestx} and of metastases in H\&E-stained lymph node images~\cite{bejnordi2017diagnostic} manifest a blatant challenge at hand, with the anomalous tissues being barely different from the normal ones (see Figure~\ref{fig:datasets}). 
\green{Although deep learning has proved useful for a variety of biomedical tasks}~\cite{nguyen2019prediction,ronneberger2015u,rajpurkar2017chexnet,kim2019deep},
\green{o}nly recently, a few groups started dedicating their effort to \green{the anomaly detection} problem~\cite{schlegl2017unsupervised,ouardini2019towards,baur2020scale}. However, to the best of our knowledge, a thorough comparison of the state-of-the-art (SOTA) solutions in the medical domain is still missing despite the pressing demand and the prospective clinical value.

In our paper, we evaluate and compare the strongest SOTA approaches  (\cite{schlegl2017unsupervised}, \cite{tuluptceva2019perceptual}, and \cite{ouardini2019towards}) on the two aforementioned medical imaging tasks. 
We find these methods either to struggle detecting such types of abnormalities, or to require a lot of time and resources for training. Moreover, the SOTA approaches 
lack a robust way of setting up model hyperparameters on \textit{new} datasets, which complicates their use in the medical domain. Thus, we revisit the problem of image anomaly detection and introduce a new 
powerful 
approach, capable of tackling these challenges. 
The proposed method leverages the efficacy of \textit{autoencoders} for anomaly detection~\cite{hinton2006reducing}, the expressiveness of \textit{perceptual loss}~\cite{johnson2016perceptual} for understanding the content in the images, and the 
\yellow{capabilities}
of the \textit{progressive growth}~\cite{karras2017progressive} to 
\yellow{approach}
training on high-dimensional image data.

Recent related studies~\cite{johnson2016perceptual,gatys2016image,zhang2018unreasonable} showed the effectiveness of deep features as a perceptual metric between images (the perceptual loss), and as a score of anomaly~\cite{tuluptceva2019perceptual}. Also, the use of the perceptual loss for training autoencoders has been very popular in a variety of tasks \cite{johnson2016perceptual,tuluptceva2019perceptual,chan2019everybody,zhang2018unreasonable,huang2018multimodal,grund2020improving,larsen2016autoencoding}
except -- inexplicably-- in the task of image anomaly detection where it has been somewhat dismissed so far. 
Trained only on normal data, the autoencoders tend to produce a high reconstruction error between the input and the output
when the input is an abnormal sample. That property has been used intensively for anomaly detection~\cite{zhai2016deep,gong2019memorizing,zhou2017anomaly,chong2017abnormal,perera2019ocgan}.
Yet, we propose to compel the autoencoder to reconstruct \textit{perceptive} or \textit{content} information of the normal images, by using \textit{only} the perceptual loss during autoencoder training. As such, the reconstructed image may not be an image altogether, but a tensor that stores the ``content'' of the original image. The main idea behind it is not to force the network to reconstruct a realistically looking image, but to let it be flexible in understanding the content of the normal data.
Section~\ref{section:perc_ae} covers the details. 

To further improve the expressiveness of the autoencoder and to allow it to capture even the fine details in the data, we propose to train the model using \yellow{the} progressive growing technique~\cite{karras2017progressive,heljakka2018pioneer}, starting from a low-resolution network and adding new layers to gradually introduce additional details during the training. In particular, we present how to achieve a smooth growth of perceptual information in the loss function, and show that this greatly improves the quality of anomaly detection in the high-resolution medical data. We will describe it in Section~\ref{section:pg}. 

Lastly, 
we propose a new approach to the basic setup of
anomaly detection model. Most approaches~\cite{perera2019ocgan,tuluptceva2019perceptual,golan2018deep,ruff2018deep,zhai2016deep}
prescribe not to use any anomaly examples during the model setup, dismissing the questions of optimization and of hyperparameter selection for such models. 
However, in reality, some types of abnormalities to detect are actually known (for example, the most frequent pathologies on the chest X-rays). Therefore, we 
consider the \textit{weakly-supervised} scenario
where a low number of anomalies with confined variability are available 
for use in optimal model hyperparameter selection
(Section~\ref{section:hyperparam}). We believe this 
scenario
reflects the real tasks encountered in practice, provides a clear pipeline for setting up the model on new data, and helps to obtain reproducible results.

To summarize our main results quantitatively, 
the proposed solution achieves 
93.4 
ROC AUC\footnote{
Area Under the Curve of Receiver Operating Characteristic in \%.} 
in the detection of metastases in H\&E stained images of lymph nodes on Camelyon16 dataset~\cite{bejnordi2017diagnostic}, and 
92.6 
in the detection of abnormal chest X-rays on the subset of NIH dataset~\cite{wang2017chestx}, which outperforms SOTA methods 
\yellow{(by 2.8\% and 5.2\% in absolute value, respectively)}.

\subsection{Contributions}
\begin{enumerate}
    \item We compare \yellow{the} three strongest SOTA anomaly detection methods (\yellow{the} hyperparameters of which we fine-tuned to their optima) in two challenging medical tasks: Chest X-rays and H\&E-stained histological images. 
    To the best of our knowledge, this is the first candid comparison 
    of anomaly detection models in digital pathology
    and in one of the largest Chest X-ray datasets available in the community\footnote{The only prior work~\cite{tang2019deep} considered a portion of dataset~\cite{wang2017chestx}.}~\cite{wang2017chestx}. We also disclose the source code of all our experiments to facilitate the development of anomaly detection in medical imaging\footnote{\url{https://github.com/ninatu/anomaly_detection/}}.
    
    \item We introduce a new anomaly detection approach that utilizes the autoencoder with the perceptual loss. The proposed model is very easy to implement and train on new data, and it provides a strong anomaly detection baseline. We further extend the proposed method with progressive growing training (in particular, we introduce how to gradually grow the perceptual information in the loss function), allowing us to adapt the anomaly detection to the \textit{high-resolution} medical data. The proposed solution outperforms SOTA methods on both datasets.
    
    \item We revisit the training setup of the anomaly detection problem statement. To address the problem of choosing the model hyperparameters in the absence of the validation dataset, we propose to relax the unrealistic assumption that no abnormal examples are available during training. We show that even a small number of \green{abnormal} images (e.g., \green{0.5\% of the training dataset}) is enough to select the hyperparameters (outcome within 2\% of the optimum). We believe that such a simple solution will standardize tuning of hyperparameters of different models, eliminate the ground for misunderstanding, and improve reproducibility.
\end{enumerate}

\section{Related Work}

\begin{figure*}[th!]
    \centering
    \includegraphics[width=0.9\textwidth]{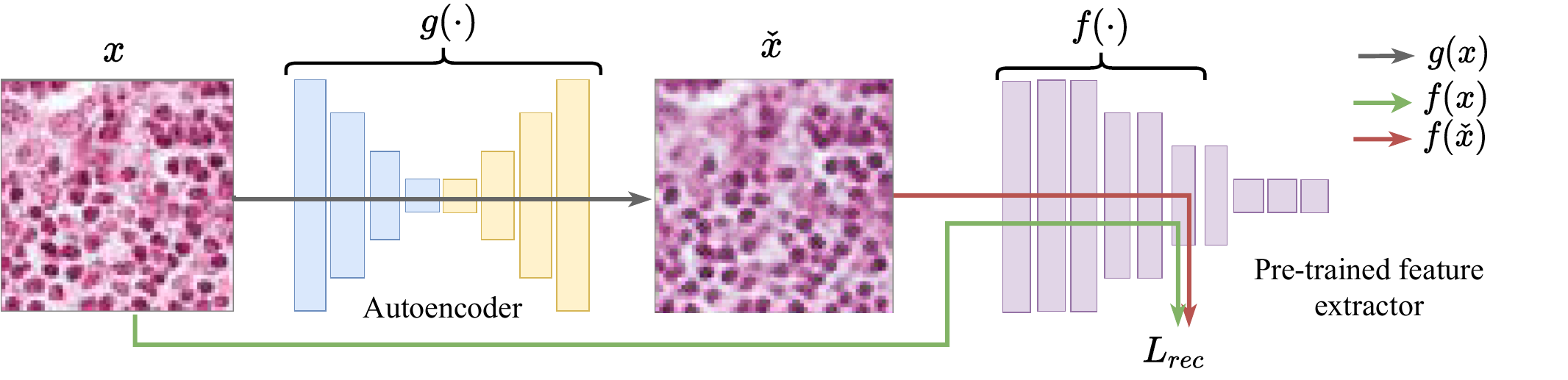}
    \caption{The proposed Deep Perceptual Autoencoder for image anomaly detection: $g$ denotes the autoencoder network, $f$ denotes a feature extractor, $x$ is an image, and $\check{x} = g(x)$ is a reconstructed ``image'' (a \textit{content} tensor). Reconstruction loss $L_{rec}$ calculates difference between deep features $f(x)$ and $f(\check{x})$.}
    \label{fig:dpa}

\end{figure*}

Anomaly detection has been extensively studied in a wide range of domains, including but not being limited to fraud detection~\cite{abdallah2016fraud}, cyber-intrusion detection~\cite{kwon2017survey}, anomalies in videos~\cite{kiran2018overview}, financial analytics~\cite{ahmed2016survey}, and the Internet of Things~\cite{mohammadi2018deep}.
An extensive survey is out of the scope of our manuscript and can be found in~\cite{chandola2009anomaly,chalapathy2019deep}. Herein, we will focus on anomaly detection in images. 

\textbf{Distribution-based methods.} Conceptually,  abnormal examples lie in low probability density areas of the ``normal'' data distribution; samples with a lower probability are thus more likely to be an anomaly.  Distribution-based methods try to predict if the new example lies in the high-probability area or not. \yellow{Kernel density estimation}~(KDE)~\cite{parzen1962estimation} or Gaussian mixture models (GMM)~\cite{mclachlan2004finite} aims to model data distribution directly. One-class SVM~\cite{chen2001one}, Isolation Forest~\cite{liu2008isolation}, SVDD~\cite{tax2004support} methods create a boundary around normal examples. The latest methods extend classical solutions by using deep data representation. For example, Deep IF~\cite{ouardini2019towards} successfully employed Isolation Forest on features extracted from a deep pre-trained network. DAGMM~\cite{zong2018deep} proposed to use GMM on learned data representation. Deep SVDD~\cite{ruff2018deep} trains a network representation to minimize the volume of a hypersphere of the normal samples. However, the most critical part of such approaches is given in learning discriminative data representation. 
As shown in~\cite{ouardini2019towards} anomaly detection performance may drop if there is a domain shift between the source dataset (for training data representation) and the target task.

\textbf{Reconstruction-based methods.} PCA and autoencoder-based~\cite{williams2002comparative} methods rely on the fact that the model trained only on normal data can not accurately reconstruct anomalies. Methods that use Generative Adversarial Networks (GANs), such as AnoGAN~\cite{schlegl2017unsupervised}, use a similar idea: the generator, trained only on normal data, cannot generate abnormal images. The reconstruction error, thus, indicates the abnormalities.  The latest methods broadly extend this idea by employing different combinations of autoencoders and adversarial losses of GAN's (OCGAN~\cite{perera2019ocgan}, GANomaly~\cite{akcay2018ganomaly}, ALOCC~\cite{sabokrou2018adversarially}, DAOL~\cite{tang2019deep} PIAD~\cite{tuluptceva2019perceptual}),  variational or robust autoencoders~\cite{zhou2017anomaly}, energy-based models (DSEBM~\cite{zhai2016deep}),  probabilistic interpretation of the latent space~\cite{abati2019latent, an2015variational}, bi-directional GANs\cite{zenati2018adversarially}, memory blocks~\cite{gong2019memorizing}, etc.  
The main difficulties of such approaches are: choosing an effective dissimilarity metric and searching for the right degree of compression (the size of the bottleneck). The~\yellow{work}~\cite{zhang2018unreasonable} shows the extraordinary effectiveness of deep features as a perceptual dissimilarity metric; however, the very perceptual loss term was not considered in the anomaly detection problem. To the best of our knowledge, only~\cite{tuluptceva2019perceptual} demonstrated a successful use of a perceptual metric for the anomaly detection task. We believe that a powerful dissimilarity measure is the key component of reconstruction-based methods. Below, we show that a \yellow{simple, yet effective,} combination of a deep autoencoder with the perceptual loss 
yields a 
\yellow{suitably accurate}
anomaly detection baseline.

A recent model Deep GEO~\cite{golan2018deep} employed a new method of image anomaly detection based on the idea of the self-supervised learning. The authors proposed to create a self-labeled dataset by applying different geometric transformations to images, with each geometric transformation (90$^{\circ}$ rotation, 180$^{\circ}$ rotation, etc.) being a new class in the dataset. After training a classifier on such a self-labeled dataset, the authors proposed to connect the abnormality of a new input to the average quality of the classification. Powerful in the natural image domain, this method, obviously, is sub-optimal for the homogeneously looking biomedical images (the rotation of which is pointless).

Despite a large number of anomaly detection methods that appeared in the recent years, only several papers~\cite{chen2018unsupervised,ouardini2019towards,tang2019deep,schlegl2017unsupervised} included medical images in their experiments, still dismissing a detailed comparison with the latest strongest models. 
An accompanying problem here is the absence of standardized benchmark for the medical anomaly detection challenge. 
Herein, we fill these gaps by implementing major SOTA methods and by comparing their performance on two popular medical datasets with different types of abnormalities.
\section{Method}

\begin{figure*}[tbh]
    \includegraphics[width=1.00\textwidth]{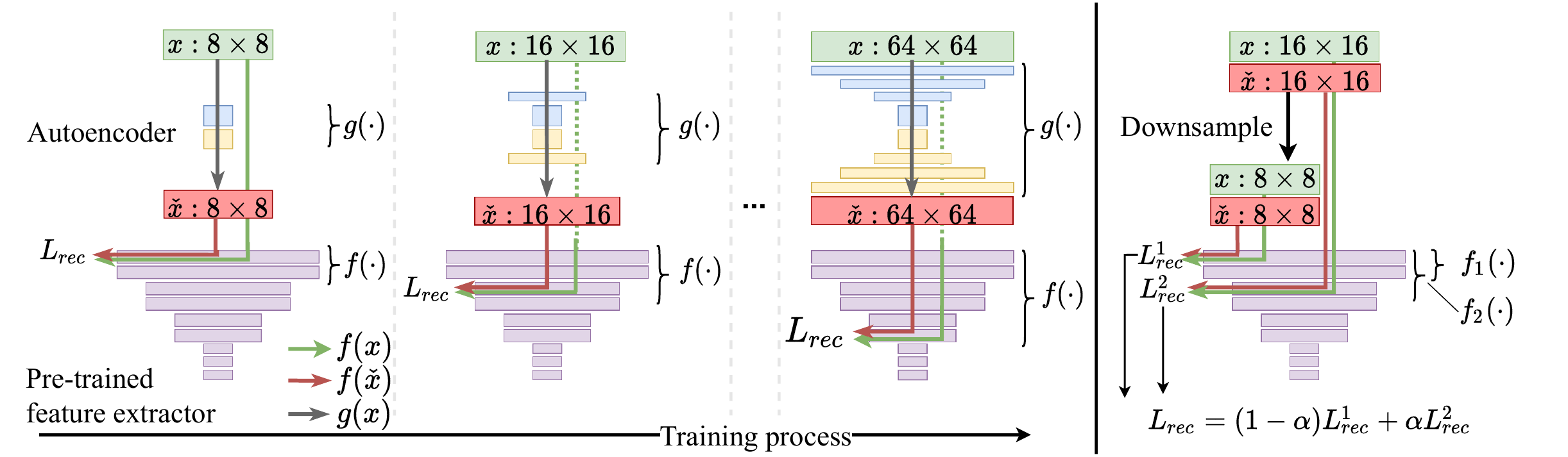}
    \caption{Progressive training process for high-resolution images. (Left) The layers are incrementally faded to the autoencoder $g$, and the depth of the features $f$ increases synchronously. (Right) The corresponding gradual increase of the ``resolution'' of the perceptual loss $L_{rec}$.}
    \label{fig:pg}
\end{figure*}

\subsection{Deep Perceptual Autoencoder}
\label{section:perc_ae}

Autoencoder-based approaches rely on the fact that autoencoders can learn shared patterns of the normal images and, then, restore them correctly.
The key idea of our method is to simplify the learning of these common factors inherent to the data, by providing a loss function that measures 
\yellow{``content dissimilarity''}
of the input and the output
\yellow{(the dissimilarity between the overall spatial structures present in the two images, without comparing the exact pixel values)}, 
which we called Deep Perceptual Autoencoder (DPA) (see Figure~\ref{fig:dpa}).
It was shown that the perceptual loss -- which computes a distance between the deep features obtained from an object classification neural network pre-trained on a large diverse dataset -- can capture the content dissimilarity of the images \cite{johnson2016perceptual,gatys2016image}. 
We further propose to use \textit{nothing but} the perceptual loss to train the autoencoder and to compute the restoration error during the evaluation, without worrying about reconstructing the \textit{whole} input information in the image by some other loss terms
(e.g., no need for the output ``image'' to look realistic).
We 
\yellow{demonstrate}
that such a loss makes the autoencoder more flexible in gaining meaningful cues of the ``normality'' of the data, ultimately leading to much better anomaly detection results.

Let $g$ be the autoencoder network, and $x$ be an image. During the training, the autoencoder minimizes the difference between $x$ and the reconstructed ``image'' $\check{x} = g(x)$, being called the reconstruction loss $L_{rec}(x, \check{x})$. To compute the perceptual loss as the reconstruction loss between $x$ and $\check{x}$, we compute the difference between the deep features of these images ($f(x)$ and $f(\check{x})$, respectively). We adopt relative-perceptual-L1 loss from Ref.~\cite{tuluptceva2019perceptual} as it is robust to noise and to changes in the image contrast\green{:} 
\begin{equation}
    L_{rec}(x, \check{x}) = \frac{\Vert\hat{f}(x) - \hat{f}(\check{x})\Vert_1}{\Vert\hat{f}(x)\Vert_1},
\end{equation}
\green{where}
\begin{equation}
\hat{f}(x) = \frac{f(x) - \mu}{\sigma}
\end{equation}
\green{are} the normalized features, with the mean $\mu$ and the standard deviation $\sigma$ of the filter responses of the layer being pre-calculated on a sufficiently large dataset.
During the evaluation, the same $L_{rec}(x, g(x))$ is used to predict the abnormality for the new input $x$.

\noindent\textbf{Remark.} Deep Perceptual Autoencoder stems from the idea of PIAD~\cite{tuluptceva2019perceptual} in that it relies on a strong image similarity metric. However, PIAD employs GANs for mapping the image distribution to the latent distribution and for the inverse mapping, which entails the following disadvantages: 
1) GANs are hard to train (training is unstable and highly sensitive to hyperparameters); 
2) GANs are time-consuming and resource-hungry (requiring alternative optimization of the generator and the discriminator); 
3) GANs require additional hyperparameter tuning due to multi-objective loss function (the adversarial and the reconstruction loss terms).
\yellow{On the contrary},
we propose the use of autoencoders that are much simpler to set up and train, 
\yellow{while combining it with perceptual loss was shown to be}
more powerful than 
GANs \yellow{utilizing} the perceptual loss. 

PIAD compels the latent vectors to be normally distributed and the reconstructed images to replicate the input images, 
\yellow{while}
Deep Perceptual Autoencoders purposely discard these restrictions,
\yellow{so that} the reconstructed image is a tensor that actually stores the ``content'' of the input image. We believe that this makes the model more flexible in \yellow{capturing key features responsible for determining the degree of} ``normality'' of the data.

\subsection{Progressive Growing}
\label{section:pg}

To improve the expressive power of the autoencoder,
we propose to train it by harnessing the methodology of progressive growth~\cite{karras2017progressive}. Illustrated in Figure~\ref{fig:pg}, the suggested pipeline gradually \textit{grows} the ``level'' of the ``perceptual'' information in the loss function. In the beginning of the training, the loss function computes the dissimilarity between the low-resolution images using the features from the coarse layers of the network,
\yellow{but}
as the training advances, the ``level'' of 
\yellow{abstraction}
\yellow{is increased through including}
deeper features.
It seems intuitively essential because the ``content" information is absent in the low-resolution images, with only the main color and the high-level structure being stored there. The novelty that we propose in our solution, therefore, is to \emph{synchronize} addition of the new layers to the autoencoder with the gradual increase of the depth of the features entailed in the calculation of the perceptual loss (see Figure~\ref{fig:pg}~(Right)).

Both the autoencoder $g$ and the perceptual loss $L_{rec}$ have a low ``resolution'' in the beginning (Figure~\ref{fig:pg}~(Left)). For example, the input and the output of the autoencoder are $8\times 8$-pixel images $x$ and $\check{x}$, and the loss $L_{rec}$ computes the distance between the features $f(x)$ and $f(\check{x})$ of the coarse layer $f$ (the pre-trained feature extractor network). As the training advances, the layers are incrementally added to the autoencoder $g$, and the depth of the features $f$ is increased.

While doubling the resolution of the autoencoder, 
for example, from $8 \times 8$ to $16 \times 16$, the new layers are 
introduced smoothly, with the parameter $\alpha$ linearly increasing from 0 to 1.
As it was proposed in~\cite{karras2017progressive,heljakka2018pioneer}, during this process, both the input $x$ and the output $\check{x}$ are the mixtures of the new high-resolution $16 \times 16$ image and the previous low-resolution $8 \times 8$ image, upsampled by a factor of two (not shown in Figure).
In a similar manner, we smoothly increase the ``level'' of information supplied to $L_{rec}$ from the features $f_1$ to the features $f_{2}$\green{:} 
\begin{multline}
    L_{rec} = \alpha * L_{rec} \big( f_{2} (x), f_{2}(\check{x})\big)  \\
    + (1 - \alpha) * L_{rec}\big( f_1(\textrm{down}(x)), f_1(\textrm{down}(\check{x})) \big),
\end{multline}
\green{where} down($\cdot$) performs downsampling by a factor of two.

Thus, the training process consists of alternating the two routines: fixed-resolution training and resolution doubling. The training starts with a small autoencoder (with most of the layers excluded) on low-resolution images with the perceptual loss over the coarse features. 
\yellow{Then} we perform the ``resolution doubling'':  we smoothly add new layers, scale up the resolution, and increase the depth of the features in the perceptual loss. Such an alternation is then repeated until the target image resolution is reached. Further details can be found in the released source code.
\subsection{Weakly-supervised Paradigm}
\label{section:hyperparam}
Any anomaly detection model has many hyperparameters,
the tuning of which is essential for the quality of the detection (in our method, these are the number of convolutions in the autoencoder, the size of the bottleneck, etc.). The majority of the anomaly detection papers declare no need to see the abnormal examples to set up their models, remaining vague with regard to how to choose the hyperparameters and how to deal with those cases when some new data needs to be analyzed by the same model. 
Some works mention tuning hyperparameters based on an unsupervised metric, like the value of the restoration error in the reconstruction-based methods~\cite{ouardini2019towards,perera2019ocgan}. 
However, lower reconstruction loss does not mean better anomaly detection quality. 
For example, better reconstruction due to a larger bottleneck can cause the autoencoder to reconstruct anomalous data accurately as well.

In practice, however, one can have access to some labeled anomalies during the model setup.  The number of such examples may be small, and they may not represent all possible abnormalities in the data, so it is typically tricky to use them in training. 
In our work, we propose 
a new \textit{weakly-supervised} training paradigm 
where a low number of labeled anomalous examples of a limited variation (i.e., a confined number of the types of anomalies) is available during the model setup as a ``validation'' or an ``optimization'' set.

This small set serves a single purpose -- select the model's hyperparameters during its setup.
Unlike works~\cite{ouardini2019towards,tang2019deep} where a small subset of \textit{all} anomalous data aims to improve the target performance, we propose to use a small subset of  \textit{limited} types of anomalies merely for the initiation of the model. This is a key difference because, in practice, it is difficult to cover all types of anomalies, even if just several examples of each. 
Below, we report extensive evaluation of how many types and examples of abnormalities are needed to validate the model. We show that even 20 abnormal images (\green{which is less than 0.5\% of the training set}) are enough to select the model within 2\% of the optimal one. 
The proposed paradigm reflects real-world clinical scenarios, allows candid comparison with standardized design of experiments, and provides a framework to \yellow{ensure} that the results are reproducible.

\section{Experiments}

\begin{table*}[ht]
\begin{center}
\begin{tabular}{ccccccccccc }
\toprule
 & AnoGAN & GANomaly & DAGMM & DSEBM & DeepSVDD & OCGAN & DeepGEO & PIAD & Deep IF & Ours (w/o p. g.)\\
\toprule
CIFAR10 & 57.6/- & 58.1/- & 57.5/- & 58.8/- & 64.8/-  & 65.7/- & 86.6/86.5 & 78.8/81.3 & 87.2/\textbf{87.3} & 83.9\\
SVHN & 53.3/- & - & 51.8/- & 57.1/- & 57.3/-  & - & 93.3/\textbf{93.5} & 77.0/76.3 & 59.0/62.4 & 80.3\\
\bottomrule
\end{tabular}
\end{center}
\caption{ROC AUC in \% for CIFAR10 and SVHN datasets averaged over all ten experiments in the dataset (see Section~\ref{section:natural}) and over three different runs per experiment (each experiment we repeated three times with different model initialization). For methods results are reported in two options: ROC AUC obtained with authors' default hyperparameters~(left), ROC AUC obtained with hyperparameters found by cross-validation in weakly-supervised paradigm~(right).}
\label{table:natural}
\end{table*}

\subsection{Datasets and Evaluation Protocol} 

We evaluated all approaches and SOTA baselines in the problem statement of \textit{novelty detection}, where the training data are assumed to be free of anomalies.

\subsubsection{Medical Images}

To perform an extensive evaluation of anomaly detection methods in the medical domain, we examined two challenging medical problems with different image characteristics and the appearance of abnormalities. 

\paragraph{Metastases Detection in Digital Pathology} Detecting metastases of lymph nodes is an extremely important variable in the diagnosis of breast cancer. However, the examination process is time-consuming and challenging. Figure~\ref{fig:datasets} shows examples of the tumor and normal tissues. Tissues \yellow{exhibiting} metastasis may differ from healthy \yellow{types} only by texture, spatial structure, or distribution of nuclei, and can be easily confused with normal tissue. We considered the task of detecting metastases in H\&E stained images of lymph nodes in the Camelyon16 challenge~\cite{bejnordi2017diagnostic}. We trained anomaly detection models only on healthy tissue aiming to identify tissue with metastases. The training dataset of Camelyon16 consists of 110 
\yellow{tumour-containing whole slide images (WSIs) and 160 normal WSIs, while the testing dataset has 80 regular WSIs and 50 tumorigenic WSIs.}

For all slides, we performed the following preprocessing. Firstly, we divided tissue from the background by applying  Otsu's thresholding~\cite{otsu1979threshold}. Then we randomly sampled  768x768 tiles (maximum 50 from one slide without overlapping) of healthy tissue (from entirely normal images) and tumor tissue (from slides with metastases) and performed color normalization~\cite{vahadane2016structure}. For the hyperparameter search we sampled tiles only from 4 out of 110 train tumor slides (validation set of confined variability).
We obtained 7612 normal training images, 200 tumor images for
validation,
and 4000~(normal) + 817~(tumor) images for the test.
During training, we randomly sampled 256x256 crops from 768x768 normalized tiles, and to the test, we used only a central 256x256 crop (to reduce border effect during normalization). The original WSIs were done with 40x magnification of tissue, but during the hyperparameter search, we also considered x10 and x20 magnification by bilinear downsampling images (256x256 to 128x128, and 64x64). 

\paragraph{Anomaly Detection on Chest X-Rays} Chest X-ray is one of the most common examinations for diagnosing various lung diseases. We considered the task of the recognition of fourteen findings, such as Atelectasis or Cardiomegaly, on the chest X-rays in the NIH dataset (ChestX-ray14 dataset)~\cite{wang2017chestx} (Figure~\ref{fig:datasets}). Searching abnormalities on a chest x-ray is challenging even for an experienced radiologist since abnormality may occupy only a small region of lungs, or be almost invisible.  The dataset consists of 112,120 frontal-view images of 30,805 unique patients: 86523 for training, 25595 for evaluation. We split the dataset into two sub-datasets having only posteroanterior~(PA) or anteroposterior~(AP) projections, because organs on them look differently. We tried different preprocessing during the hyperparameter search: rescaling to 256x256, 128x128, and 64x64 and histogram equalization, central crop (3/4 of the image size) to delete ``noisy'' \yellow{borders}.  We considered images without any disease marker as ``normal'' and used them for training. 
\yellow{Abnormal images for hyperparameter searching comprised of}
the training images of the most frequent disease (`Infiltration') out of fourteen possibilities. We also evaluated model on subset containing ``clearer'' normal/abnormal cases (provided by ~\cite{tang2019deep}). This subset consists \yellow{of} 4261 normal images for training, 849 normal and 857 abnormal images for validation, and 677 normal and 677 abnormal images for testing.

\subsubsection{Natural Images}
\label{section:natural}

We also evaluate the methods on two natural image benchmarks CIFAR10~\cite{krizhevsky2009learning} and SVHN~\cite{netzer2011reading}.
Both datasets provide an official train-test split and consist of 10 classes. Following previous works~\cite{ruff2018deep,golan2018deep,perera2019ocgan,ouardini2019towards,tuluptceva2019perceptual,zhai2016deep}, we used a one-vs-all evaluation protocol: we design 10 different experiments, where only one class is alternately considered as normal, while others treated as abnormal.
In all experiments, 
\yellow{images were rescaled  to 32x32 resolution.}
During the hyperparameter search, we tried different preprocessing: \yellow{1)} using the original RGB-images and \yellow{2)} converting \yellow{the} images to grayscale. 
We randomly sampled one abnormal class of the train set as a validation set with abnormal images (that has only one type of abnormalities out of nine).  These conditions were fixed in all methods compared beneath.

\subsubsection{Metrics}
\green{
Following the convention in the field of anomaly detection }~\cite{perera2019ocgan,ouardini2019towards,golan2018deep,tuluptceva2019perceptual,ruff2018deep,akcay2018ganomaly,tang2019deep}\green{, we used the Area Under the Curve of Receiver Operating Characteristic (ROC AUC) as an evaluation metric that integrates the classification performance (normal \textit{vs.} abnormal) for all decision thresholds. This precludes from the need to choose a threshold for the predicted abnormality scores, allowing for assessing the performance of the models ``probabilistically'' and without a bias.
}

\subsection{Baselines.}

We considered the following strongest SOTA baselines of different paradigms: Deep GEO~\cite{golan2018deep} 
Deep IF~\cite{ouardini2019towards} 
and PIAD~\cite{tuluptceva2019perceptual}.
On natural images we also competed against AnoGAN~\cite{schlegl2017unsupervised}, GANomaly~\cite{akcay2018ganomaly}, DAGMM~\cite{zong2018deep}, DSEBM~\cite{zhai2016deep}, DeepSVDD~\cite{ruff2018deep}, and OCGAN~\cite{perera2019ocgan} methods.
On the NIH dataset, \yellow{we} also compared our results to DAOL framework~\cite{goodfellow2014generative,tang2019deep}, purposely developed for detecting anomalies in chest X-rays. 

\subsection{Implementation details.} 

We implemented Deep IF and PIAD approaches using extensive descriptions provided by \yellow{the} authors. For GANomaly and Deep GEO, we adapted the official code for our experiment setups. Results of DAOL and OCGAN methods were obtained in the corresponding papers. For other approaches, we used results as reported in~\cite{ouardini2019towards}.
For the strongest baselines, we also perform a hyperparameter tuning precisely as it is proposed herein (for a standardized comparison). For the Deep GEO approach, we searched for an optimal number of the classifier's training epochs (we find the method to be sensitive to this parameter). For Deep IF, we searched for the best feature representation -- the best layer of the feature extractor network. For PIAD, we searched for the optimal size of the latent vector, the best feature layer in the relative-perceptual-L1 loss, and the \yellow{most suitable} number of training epochs. Also, for all algorithms, we searched for the best image preprocessing to yield the highest scores. 

For the proposed approach, in all experiments, we used autoencoders with pre-activation residual blocks.
For the computation of the relative-perceptual-L1 loss, we used the VGG19 network that was pre-trained on \textit{ImageNet}. We trained the autoencoder until after the loss on the hold-out set of normal images stops decreasing. During the hyperparameter selection, we search for the best size of the autoencoder bottleneck and for the best feature layer of relative-perceptual-L1 loss. Further details \yellow{are} covered in the released algorithm code.

Hyperparameter search was performed by maximizing average ROC AUC over 3 ``folds''. Namely, we split training normal data into 3 subsets (as a 3-fold cross-validation) and did the same for validating the abnormal data. In each run, we left one normal subset and one abnormal subset for testing, and used the other two normal subsets for training.
Only in the experiments with the NIH subset, Ref.~\cite{tang2019deep}, we did not perform cross-validation but \yellow{ran the experiment thrice} on the same train-validation split to precisely replicate the DAOL experimental settings. 

\subsection{Results}


\begin{table*}[ht]
\begin{center}
\resizebox{1\linewidth}{!}{ 
\begin{tabular}{ ccccccc }
\toprule
 & DAOL & Deep GEO & PIAD & Deep IF & Ours (w/o p. g.) &  Ours (with p. g.)\\
\toprule
Camelyon16 & - & $52.4\pm11.1$/$45.9\pm2.1$ & 
$85.4\pm2.0$/$89.5\pm0.6$ &  
$87.6\pm1.5$/$90.6\pm0.3$ &
$92.7\pm0.4$ &
$\bm{93.4\pm0.3}$ \\

NIH (a subset) & $-/80.5\pm2.1$ & $85.8\pm0.6$/ $85.3\pm1.0$ &
$88.0\pm1.1$/$87.3\pm0.9$ & 
$76.6\pm2.7$ /$85.3\pm0.4$ & 
$92.0\pm0.2$ &
$\bm{92.6\pm0.2}$ \\

NIH (PA proj.) & - & $60.2\pm2.6$/$63.6\pm0.6$ & 
$68.0\pm0.2$/$68.7\pm0.5$ & 
$52.2\pm0.5$/$47.2\pm0.4$ & 
$70.3\pm0.2$ &
$\bm{70.8\pm0.1}$ \\

NIH (AP proj.) & - &$53.1\pm0.3$/$54.4\pm0.6$ & 
$57.4\pm0.4$/$\bm{58.6\pm0.3}$ & 
$54.3\pm0.5$/$56.1\pm0.2$ & 
$\bm{58.6\pm0.1}$ &
  $58.5\pm0.0$ \\
\bottomrule
\end{tabular}
}
\end{center}
\caption{
ROC AUC in \% with standard deviation (over 3 runs). For baselines results are reported in two options: ROC AUC obtained with authors' default hyperparameters~(left), ROC AUC obtained with hyperparameters found by cross-validation in weakly-supervised paradigm~(right). For our method, results are showed with and without progressive growing regime of training. }
\label{table:main}
\end{table*}
\begin{table}[t]
\begin{center}
\resizebox{1\linewidth}{!}{ 
\begin{tabular}{cccc }
\toprule
 & PIAD & Ours (w/o p. g.) & Ours (with p. g.)\\
\toprule
Camyleon16 & 84 & 105 & 160 \\
NIH (a sub.) & 287 & 59 & 177 \\
NIH (PA proj.) & 151 & 89 & 159  \\
NIH (AP proj.) & 275 & 70 & 107\\
\bottomrule
\end{tabular}
}
\end{center}
\caption{Average training time (minutes). Experiments were run on GeForce GTX 1080 Ti with Pytorch 1.4.0.}
\label{table:time}
\end{table}
\begin{figure*}[tbp]
    \centering
    \begin{subfigure}[]{0.43\textwidth}
        \centering
        \captionsetup{justification=centering, margin=0cm}
        \begin{subfigure}[]{0.33\textwidth}
            \begin{overpic}[width=1\textwidth]{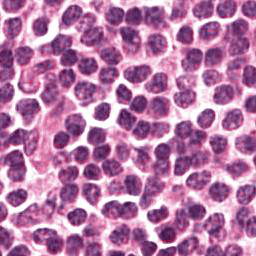}
            \put(0,2){\colorbox{white}{\textbf{0.35}}}
            \end{overpic}
        \end{subfigure}%
        \begin{subfigure}[]{0.33\textwidth}
            \begin{overpic}[width=1\textwidth]{{images/camelyon16/normal/1799_0.42}.png}
            \put(0,2){\colorbox{white}{\textbf{0.42}}}
            \end{overpic}
        \end{subfigure}%
        \begin{subfigure}[]{0.33\textwidth}
            \begin{overpic}[width=1\textwidth]{{images/camelyon16/normal/1238_0.45}.png}
            \put(0,2){\colorbox{white}{\textbf{0.45}}}
            \end{overpic}
        \end{subfigure}%
    \end{subfigure}%
    \hspace{0.05\textwidth}
    \begin{subfigure}[]{0.43\textwidth}
        \centering
        \captionsetup{justification=centering, margin=0cm}
        \begin{subfigure}[]{0.33\textwidth}
            \begin{overpic}[width=1\textwidth]{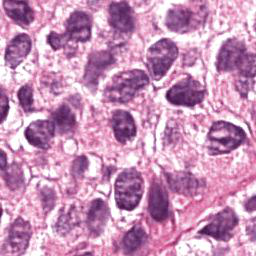}
            \put(0,2){\colorbox{white}{\textbf{0.54}}}
            \end{overpic}
        \end{subfigure}%
        \begin{subfigure}[]{0.33\textwidth}
            \begin{overpic}[width=1\textwidth]{{images/camelyon16/anomaly/464_0.52}.png}
            \put(0,2){\colorbox{white}{\textbf{0.52}}}
            \end{overpic}
        \end{subfigure}%
        \begin{subfigure}[]{0.33\textwidth}
            \begin{overpic}[width=1\textwidth]{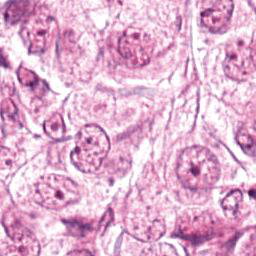}
            \put(0,2){\colorbox{white}{\textbf{0.55}}}
            \end{overpic}
        \end{subfigure}%
    \end{subfigure}
    \begin{subfigure}[]{0.43\textwidth}
        \centering
        \captionsetup{justification=centering, margin=0cm}
        \begin{subfigure}[]{0.33\textwidth}
            \begin{overpic}[width=1\textwidth]{{images/nih/normal/556_0.29}.png}
            \put(0,2){\colorbox{white}{\textbf{0.29}}}
            \end{overpic}
        \end{subfigure}%
        \begin{subfigure}[]{0.33\textwidth}
            \begin{overpic}[width=1\textwidth]{{images/nih/normal/562_0.39}.png}
            \put(0,2){\colorbox{white}{\textbf{0.39}}} 
            \end{overpic}
        \end{subfigure}%
        \begin{subfigure}[]{0.33\textwidth}
            \begin{overpic}[width=1\textwidth]{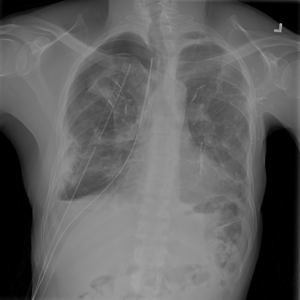}
            \put(0,2){\colorbox{white}{\textbf{0.30}}}
            \end{overpic}
        \end{subfigure}%
        \caption*{\small Normal}
    \end{subfigure}%
    \hspace{0.05\textwidth}
    \begin{subfigure}[]{0.43\textwidth}
        \centering
        \captionsetup{justification=centering, margin=0cm}
        \begin{subfigure}[]{0.33\textwidth}
            \begin{overpic}[width=1\textwidth]{{images/nih/anomaly/573_0.60}.png}
            \put(0,2){\colorbox{white}{\textbf{0.60}}}
            \end{overpic}
        \end{subfigure}%
        \begin{subfigure}[]{0.33\textwidth}
            \begin{overpic}[width=1\textwidth]{{images/nih/anomaly/402_0.49}.png}
            \put(0,2){\colorbox{white}{\textbf{0.49}}}
            \end{overpic}
        \end{subfigure}%
        \begin{subfigure}[]{0.33\textwidth}
            \begin{overpic}[width=1\textwidth]{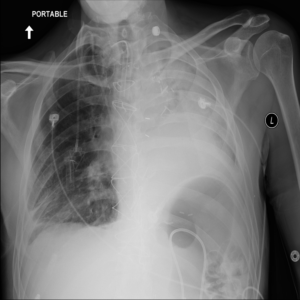}
            \put(0,2){\colorbox{white}{\textbf{0.66}}}
            \end{overpic}
        \end{subfigure}%
        \caption*{\small Anomaly}
    \end{subfigure}
    \caption{
    Examples of normal (left) and anomaly (right) images of H\&E-stained lymph node of Camelyon16 challenge~\cite{bejnordi2017diagnostic} (top) and chest X-rays of NIH dataset~\cite{wang2017chestx} (bottom). We also showed the predicted anomaly score by the proposed method. The higher the score, the more likely to be an anomaly. Notice how the proposed method spots even the borderline cases.
    }
    \label{fig:predictions}
\end{figure*}

\subsubsection{Natural Images}

As mentioned above, for CIFAR10 and SVHN datasets, we conducted ten experiments, where each class alternatively was considered normal. 
In such experiments, an anomaly is an image of an object of a different class. Therefore, abnormal images are very different from normal data (compared to anomalies on medical images), but normal data also have high variability. 
The average results over all experiments in a dataset are reported in Table~\ref{table:natural}. 
Notice, while testing on these datasets, we do not use progressive growth in our method because the image resolution is only $32\times32$.

The approaches that we called the strongest baselines (Deep GEO, PIAD, Deep IF) and our method significantly outperform other methods, with margin 20\% (except for Deep IF on SVHN dataset). 
The Deep GEO approach, which classifies the geometric transformations of images, excels in distinguishing digits from each other (SVHN dataset). The reason for that is that digits have a simple geometrical structure, and their geometric transformations are easily distinguishable. Our approach shows the second-best result. However, Deep IF fails -- features obtained by \yellow{the \textit{ImageNet}-pre-trained} neural network turned out to be not discriminative for this task. 

Images of CIFAR10 dataset have a more challenging geometrical structure than SVHN ones, \yellow{hence} Deep GEO shows lower performance. However, since the domain shift between \textit{ImageNet} and CIFAR10 dataset is smaller, Deep IF also shows good results. Our 
\yellow{proposed method closely approaches the reported performance of}
the leaders Deep GEO and Deep IF.  We noticed that our approach in both datasets outperformed PIAD by $\sim$3\%.

We report that reconstruction-based approaches, like ours and PIAD, are inferior to Deep GEO for the natural image tasks. We hypothesize \yellow{that} this stems from the high variability of the normal images, and the autoencoder overgeneralizes on the anomaly data. Indeed, during hyperparameter tuning, we search for the optimal autoencoder capacity, where the autoencoder reconstructs the normal data well but does not generalize on the other data. However, when the training data \yellow{is} highly variable, the autoencoder generalizes better on the unseen classes.

\subsubsection{Medical Images}

\begin{table*}[ht] 
\begin{center}
\begin{tabular}{lcccc}
\toprule
 & Camelyon16 & NIH (a subset) & NIH (PA proj.) & NIH (AP proj.)\\
\toprule
(1) L1 + unsupervised & $21.1\pm1.4$ & $70.8\pm0.6$ & $66.5\pm0.1$ & $52.4\pm0.1$\\
(2) PL + unsupervised & $87.9\pm0.6$ & $89.3\pm0.2$ & $68.9\pm0.1$ & $56.4\pm0.2$\\
(3) PL + weakly-supervised & $92.7\pm0.4$ & $92.0\pm0.2$ & $70.3\pm0.2$ & $\bm{58.6\pm0.1}$\\
\cmidrule{1-5}
(4) PL + $1 \cdot$ adv + weakly-supervised & $79.4\pm4.0$ & $64.4\pm7.8$ & $52.3\pm3.3$ & $51.5\pm3.4$\\
(5) PL + $0.1 \cdot$ adv + weakly-supervised & $90.8\pm0.7$ & $82.2\pm2.6$ & $59.2\pm1.4$ & $55.4\pm0.9$\\
\cmidrule{1-5}
(6) PL + $1\cdot$ L1 + weakly-supervised & $75.3\pm1.6$ & $91.7\pm0.4$ & $70.7\pm0.2$ & $57.3\pm0.1$\\
(7) PL + $0.1\cdot$ L1 + weakly-supervised & $93.0\pm0.3$ & $92.0\pm0.1$ & $70.6\pm0.2$ & $58.5\pm0.1$\\
\cmidrule{1-5}
(8) PL + $1 \cdot$ L1 + $1\cdot$ adv + weakly-supervised & $57.5\pm6.3$ & $59.3\pm5.0$ & $50.1\pm2.0$ & $51.7\pm0.8$\\
(9) PL + $0.1 \cdot$ L1 + $0.1\cdot$ adv + weakly-supervised & $90.6\pm1.0$ & $78.2\pm1.0$ & $60.8\pm1.8$ & $55.5\pm0.4$\\
\cmidrule{1-5}
(10) PL + weakly-supervised + progressive growing & $\bm{93.4\pm0.3}$ & $\bm{92.6\pm0.2}$ & $\bm{70.8\pm0.1}$ & $58.5\pm0.0$\\
\bottomrule
\end{tabular}
\end{center}
\caption{Ablation study. ROC AUC in \% with standard deviation (over 3 runs).}
\label{table:ablation}
\end{table*}

\begin{figure*}[ht] 
    \centering
    \begin{subfigure}[]{0.48\textwidth}
        \captionsetup{justification=centering, margin=0cm}
        \centering
        \caption*{Camelyon16}
        \includegraphics[width=1\textwidth]{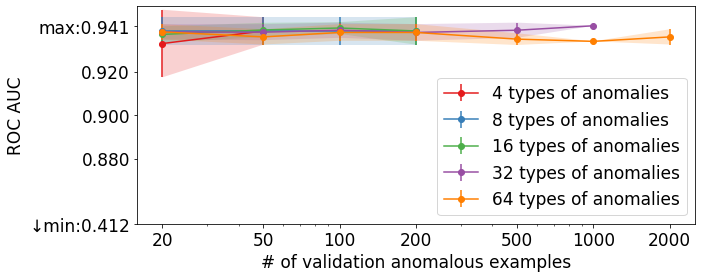}
    \end{subfigure}%
    \hfill
    \begin{subfigure}[]{0.48\textwidth}
        \captionsetup{justification=centering, margin=0cm}
        \centering
        \caption*{NIH (a subset)}
        \includegraphics[width=1\textwidth]{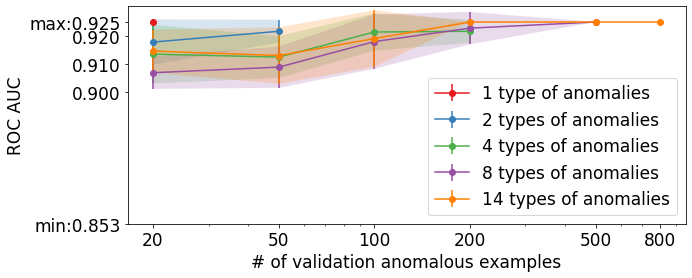}
    \end{subfigure}%
    
    \caption{Dependence of the quality of anomaly detection (of our approach) on the number of anomaly examples (the x-axis) and their variability (the different lines) in the validation set. The highest (max) and the lowest (min) performance achievable on these hyperparameter spaces are shown on the plots. For Camelyon16, we consider metastases tiles from one slide as abnormality of one type, for NIH dataset, type of abrnomality is unique finding. We used the same 3-fold cross-validation split and hyperparameter space as in previous experiments.   We sampled a validation set for each configuration (\# anomaly types, \# anomaly examples) seven times. For each sample of the validation set, we selected the best hyperparameters on the cross-validation split. Then we evaluated the quality of the model trained on all training images with chosen hyperparameters on test split. Here we showed mean and std of test ROC AUC's (computed over three samples of the validation set for this configuration)}
    \label{fig:hyperparams}
\end{figure*}

Given Deep GEO, PIAD, and Deep IF are superior to the other methods on natural image datasets, we chose them for evaluation on the medical images too, where such a diverse variability is absent (see Table~\ref{table:main}).  For our method, we report results obtained with and without the progressive growing training regime.

Remarkably, our approach significantly outperforms Deep GEO and Deep IF in \yellow{both} medical datasets. 
The Deep GEO shows poor performance on the digital pathology data \yellow{(only 52.4\% with the default  hyperparameters proposed by the authors)}, where the images are invariant to geometric transformations. Indeed, digital pathology scans do not have space orientation; rotations and translations of them are not distinguishable. 
We can tell that the Deep GEO is not applicable to such data. For NIH PA and AP projections, ROC AUC's are also very low. Our hypothesis is that if abnormality occupy a small region of the image, the classifier still distinguishes the geometric transformations well, so the quality clasification hardly indicates such abnormalities. For NIH (a subset) with more ``obvious'' abnormalities (the abnormal region is larger), the Deep GEO approach shows better results. 

\yellow{Deep IF displayed inferior performance on the NIH (PA proj.) and NIH (AP proj.) datasets, achieving ROC AUC scores of 52.2\% and 54.3\% respectively (Table}~\ref{table:main}\yellow{, default hyperparameters proposed by the authors).}
\yellow{This might probably be due to the}
domain shift between ImageNet and X-ray images, \yellow{whereby the features} obtained from the pre-trained network turned out to be not discriminative for this task. However, for the Camelyon16 and NIH (a subset) experiments, \yellow{the} ROC AUC is  quite high. We 
\yellow{postulate that if the feature extractor network was pre-trained on images which closely-resembled those of the X-ray image analysis task, Deep IF would demonstrate a significantly-improved performance on the NIH (PA proj.) and the NIH (AP proj.) datasets as well.}
We conclude that the weakest side of this approach is feature representation. \yellow{Currently,} there \yellow{exists} no explicit algorithm for 
\yellow{obtaining discriminative feature spaces for differing experimental setups.}

It is worth noting that our method also uses the features pre-trained on \textit{ImageNet}, but 
for image comparison (rather than image representation), since the learnt features (though not being
discriminative due to the domain shift), 
\yellow{) might still prove useful in determining the key differences between images}
(via the perceptual loss). 

With a smaller margin, our algorithm (in both options: with and without progressive growing) is also ahead of the PIAD method.  
We would like to highlight that our method \yellow{is generally} easier and faster to train \yellow{while being} the least resource-hungry amongst the models considered herein. 
The average time of training of \yellow{the} final models is provided in Table~\ref{table:time}. 
\yellow{Moreover, the use of the progressive growing training approach gained us an additional 1\% in the image quality (as depicted by the ROC AUC scores in Table 2).}

We illustrate the predictions of our model in Figure~\ref{fig:predictions}.

\subsection{Hyperparameters Tuning Analysis}

While proposing the use of a small and restricted set of anomalies during the model setup, we naturally asked exactly how many anomaly types and how many anomaly examples would be required. In Figure~\ref{fig:hyperparams}, we demonstrate the dependence of the quality of the anomaly detection on the number of anomaly examples and their variability in the validation set.
The experiment shows that even a small number \yellow{of} abnormal samples (for example, 20) of \textit{one type} of anomaly is enough, to ``reject'' inferior hyperparameter configurations. In the two experiments considered, having 20 abnormal examples of the same type of abnormality \green{(which is less than 0.5\% of the normal examples used in training)} proved sufficient to select the hyperparameters within the 2\% \yellow{margin of the} optimal configuration.

\subsection{Ablation study}

\begin{table*}[h!]
\caption{\green{Results on natural images. ROC AUC in \% with std in each experimental configuration on the CIFAR 10 and SVHN datasets (each experiment was repeated three times with different model initialization). We used a one-vs-all evaluation protocol: 10 different experiments for each dataset were designed, where only one class (a column name) is alternately considered as normal,  while the others are treated as abnormal. For these methods, the results are reported for two options: ROC AUC obtained with the default hyperparameters proposed by the authors of corresponding works (default), and ROC AUC obtained with the hyperparameters found by cross-validation in weakly-supervised paradigm (weakly s.))}}
\setlength\tabcolsep{2.5pt}
\begin{center}
\resizebox{1\linewidth}{!}{ 
\begin{tabular}{cccccccccccc}
\toprule
&&\multicolumn{10}{c}{CIFAR10} \\
\cmidrule{3-12}
 & hyperparams & plane & car & bird & cat & deer & dog & frog & horse & ship & truck\\
\cmidrule{2-12}
\multirow{2}{*}{Deep GEO} & default & $75.4$$\pm0.9$ & $96.0$$\pm0.2$ & $79.9$$\pm1.9$ & $73.6$$\pm0.2$ & $87.4$$\pm0.4$ & $\bm{87.6}$$\bm{\pm0.7}$ & $85.2$$\pm0.9$ & $95.1$$\pm0.1$ & $94.3$$\pm0.0$ & $91.3$$\pm0.3$\\
 & weakly-s. & $75.7$$\pm1.0$ & $96.0$$\pm0.2$ & $\bm{80.4}$$\bm{\pm1.1}$ & $72.9$$\pm0.9$ & $88.0$$\pm0.2$ & $86.3$$\pm0.9$ & $84.6$$\pm0.5$ & $\bm{95.4}$$\bm{\pm0.0}$ & $\bm{94.3}$$\bm{\pm0.2}$ & $91.4$$\pm0.5$\\
\cmidrule{2-12}
\multirow{2}{*}{PIAD} & default & $81.8$$\pm0.1$ & $87.1$$\pm0.3$ & $74.9$$\pm0.3$ & $60.7$$\pm0.2$ & $78.1$$\pm0.5$ & $70.6$$\pm1.4$ & $81.7$$\pm0.8$ & $84.4$$\pm0.4$ & $86.3$$\pm0.4$ & $82.3$$\pm0.6$\\
 & weakly-s. & $84.3$$\pm0.2$ & $86.7$$\pm1.1$ & $74.4$$\pm0.9$ & $59.6$$\pm2.1$ & $85.0$$\pm1.1$ & $73.6$$\pm1.1$ & $83.8$$\pm1.2$ & $87.0$$\pm1.1$ & $88.8$$\pm0.2$ & $89.4$$\pm0.7$\\
\cmidrule{2-12}
\multirow{2}{*}{Deep IF} & default & $85.2$$\pm1.2$ & $94.3$$\pm0.4$ & $72.5$$\pm4.0$ & $\bm{76.8}$$\bm{\pm1.2}$ & $\bm{89.9}$$\bm{\pm0.7}$ & $86.1$$\pm1.0$ & $90.3$$\pm1.7$ & $89.1$$\pm1.0$ & $92.0$$\pm1.0$ & $95.6$$\pm0.1$\\
 & weakly-s. & $\bm{87.1}$$\bm{\pm0.9}$ & $\bm{97.0}$$\bm{\pm0.3}$ & $75.2$$\pm2.9$ & $73.7$$\pm1.8$ & $88.9$$\pm1.0$ & $85.0$$\pm2.6$ & $\bm{90.5}$$\bm{\pm0.9}$ & $86.3$$\pm1.7$ & $93.4$$\pm0.3$ & $\bm{95.7}$$\bm{\pm0.3}$\\
\cmidrule{2-12}
Ours & weakly-s. & $86.5$$\pm0.2$ & $92.2$$\pm0.3$ & $76.8$$\pm0.6$ & $58.7$$\pm1.2$ & $85.1$$\pm0.4$ & $77.7$$\pm0.9$ & $88.9$$\pm0.1$ & $89.1$$\pm0.2$ & $91.4$$\pm0.5$ & $92.2$$\pm0.4$\\
\toprule
&&\multicolumn{10}{c}{SVHN} \\
\cmidrule{3-12}
 &  & 0 & 1 & 2 & 3 & 4 & 5 & 6 & 7 & 8 & 9\\
\cmidrule{2-12}
\multirow{2}{*}{Deep GEO} & default & $89.0$$\pm0.6$ & $84.1$$\pm1.1$ & $96.9$$\pm0.1$ & $\bm{91.3}$$\bm{\pm0.3}$ & $97.3$$\pm0.0$ & $96.2$$\pm0.3$ & $96.0$$\pm0.2$ & $98.2$$\pm0.1$ & $\bm{86.4}$$\bm{\pm0.3}$ & $97.4$$\pm0.1$\\
 & weakly-s. & $\bm{90.6}$$\bm{\pm0.6}$ & $\bm{84.8}$$\bm{\pm0.6}$ & $\bm{97.2}$$\bm{\pm0.2}$ & $91.1$$\pm0.1$ & $\bm{97.5}$$\bm{\pm0.1}$ & $\bm{96.3}$$\bm{\pm0.0}$ & $\bm{96.2}$$\bm{\pm0.2}$ & $\bm{98.4}$$\bm{\pm0.0}$ & $85.6$$\pm0.9$ & $\bm{97.6}$$\bm{\pm0.2}$\\
\cmidrule{2-12}
\multirow{2}{*}{PIAD} & default & $85.6$$\pm0.4$ & $79.2$$\pm1.1$ & $74.8$$\pm0.4$ & $69.2$$\pm0.0$ & $77.3$$\pm0.8$ & $74.6$$\pm0.9$ & $76.3$$\pm0.7$ & $77.5$$\pm0.4$ & $78.0$$\pm0.2$ & $77.4$$\pm0.3$\\
 & weakly-s. & $86.3$$\pm0.9$ & $80.2$$\pm0.9$ & $76.2$$\pm0.8$ & $71.4$$\pm1.1$ & $77.0$$\pm0.5$ & $71.9$$\pm0.9$ & $70.6$$\pm0.5$ & $78.4$$\pm0.2$ & $79.6$$\pm0.7$ & $71.9$$\pm0.8$\\
\cmidrule{2-12}
\multirow{2}{*}{Deep IF} & default & $65.3$$\pm1.1$ & $68.7$$\pm1.8$ & $51.9$$\pm0.8$ & $57.1$$\pm1.5$ & $56.7$$\pm2.1$ & $64.9$$\pm1.5$ & $50.9$$\pm0.8$ & $56.2$$\pm1.9$ & $63.7$$\pm0.9$ & $54.3$$\pm1.1$\\
 & weakly-s. & $75.0$$\pm1.2$ & $70.5$$\pm1.5$ & $51.1$$\pm0.8$ & $59.0$$\pm0.9$ & $57.7$$\pm1.4$ & $68.4$$\pm0.6$ & $54.5$$\pm0.2$ & $58.7$$\pm0.7$ & $69.6$$\pm1.6$ & $59.1$$\pm1.5$\\
\cmidrule{2-12}
Ours & weakly-s. & $88.4$$\pm0.2$ & $82.7$$\pm0.8$ & $80.0$$\pm0.8$ & $72.9$$\pm0.1$ & $79.1$$\pm0.7$ & $77.4$$\pm0.7$ & $78.0$$\pm0.8$ & $79.0$$\pm0.2$ & $83.5$$\pm0.2$ & $82.1$$\pm0.3$\\
\bottomrule
\end{tabular}
}
\end{center}
\label{table:naturla_by_classes}
\end{table*}

To stress the importance of every component proposed herein, we performed an extensive ablation study. Table~\ref{table:ablation} considers ten ablation scenarios.

\textbf{(1)}: Autoencoder (AE) training with the L1 loss and the hyperparameter optimization using \textit{unsupervised} criteria (the reconstruction loss).

\textbf{(2)}: The same, but with L1 replaced by perceptual loss (PL).

\textbf{(3)}: The previous one with the hyperparameters corresponding to the best validation ROC AUC (\textit{weakly-supervised} scenario). 

\textbf{(4)--(9)}: Here, we added the adversarial loss (with weights 1 and 0.1) or L1 norm, or both of them to the loss function during the training (to force the reconstructed image to have a realistic look or to restore the whole input image). 
To compute adversarial loss we trained the discriminator jointly with autoencoder using Wasserstein GAN with a Gradient Penalty objective~\cite{gulrajani2017improved}.

\textbf{(10)}: The last training scenario finally considers the progressive growing. 

Remarkably, the use of the perceptual loss~\textbf{(2)} outperforms the mere L1 norm~\textbf{(1)} with a large margin.
We also \yellow{observed} that the method of selecting the hyperparameters by revealing a subset of anomalies of confined variability \textbf{(3)} noticeably benefits the anomaly detection performance (\yellow{when} compared to the unsupervised criteria \textbf{(2)}). 
We also note the advantage of our approach over the autoencoder that encourages fully restored or realistically looking images \textbf{(4)--(9)} (using additional adversarial or L1 norm loss). 
We thus confirm our hypothesis that the use of the perceptual loss alone provides the autoencoder with more flexibility to gain a meaningful interpretation of the ``normality''. Our experiments demonstrate that the additional losses only deteriorate the performance. 
Finally, the proposed progressive growing technique \textbf{(10)} allow us to gain further improvements, solidifying the entire model as the new medical image anomaly detection baseline.

\subsection{Limitations}

The proposed method excelled on the medical datasets but was unable to outperform \yellow{the assayed SOTA on} natural image baselines. 
We, thus, hypothesize that the method may be \yellow{better suited for data having} low to medium variability \yellow{(}such as the medical images from the same image acquisition modality\yellow{)} rather than high variability datasets \yellow{(}such as CIFAR10 or SVHN\yellow{)}. \yellow{The} high diversity \yellow{present in} the natural data may lead to overgeneralization of the autoencoder: in \yellow{this case}, the autoencoder may produce low reconstruction error even for the abnormal data. A controlled study \yellow{on} the dependence of the performance of all methods on data diversity requires specifically prepared datasets (perhaps, with synthetic anomalies and a strict measure of `data diversity') and will be the subject of future work. Still, one may argue that a very sensitive anomaly detection algorithm, capable of discriminating against \yellow{similarly-looking} normal data and tuned for a given imaging modality, would actually be preferred by 
\yellow{individual imaging niches over more generalized}
but less sensitive methods.

\begin{figure*}[ht] 
    \centering
    \begin{subfigure}[]{0.33\textwidth}
        \captionsetup{justification=centering, margin=0cm}
        \centering
        \includegraphics[width=1\textwidth]{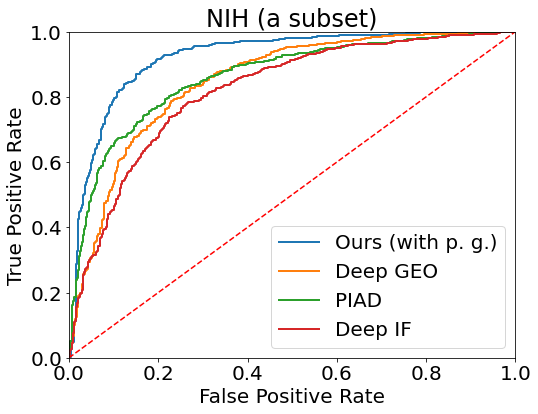}
    \end{subfigure}%
    \hfill
    \begin{subfigure}[]{0.33\textwidth}
        \captionsetup{justification=centering, margin=0cm}
        \centering
        \includegraphics[width=1\textwidth]{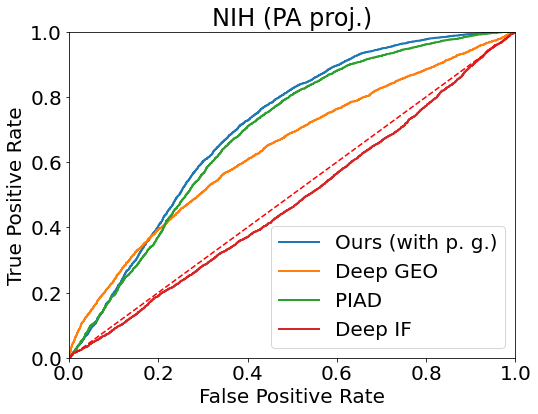}
    \end{subfigure}%
    \hfill
    \begin{subfigure}[]{0.33\textwidth}
        \captionsetup{justification=centering, margin=0cm}
        \centering
        \includegraphics[width=1\textwidth]{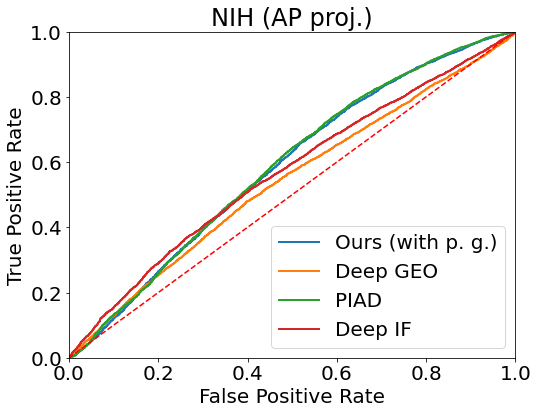}
    \end{subfigure}%
    \caption{\green{ROC curves of Deep GEO, PIAD, Deep IF, and the proposed model on the NIH (a subset), NIH (PA proj.), and NIH (AP proj.) datasets.}}
    \label{fig:roc_curves}
\end{figure*}

\begin{figure*}[ht] 
    \centering
    \begin{subfigure}[]{0.33\textwidth}
        \captionsetup{justification=centering, margin=0cm}
        \centering
        \includegraphics[width=1\textwidth]{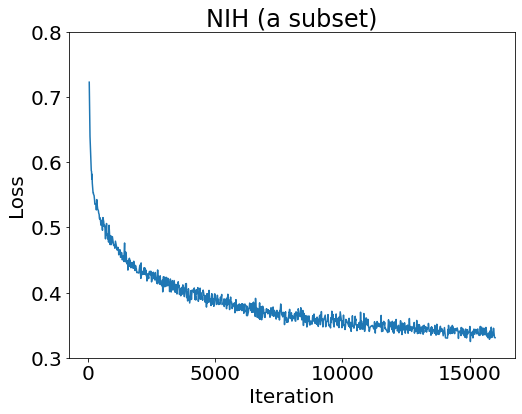}
    \end{subfigure}%
    \hfill
    \begin{subfigure}[]{0.33\textwidth}
        \captionsetup{justification=centering, margin=0cm}
        \centering
        \includegraphics[width=1\textwidth]{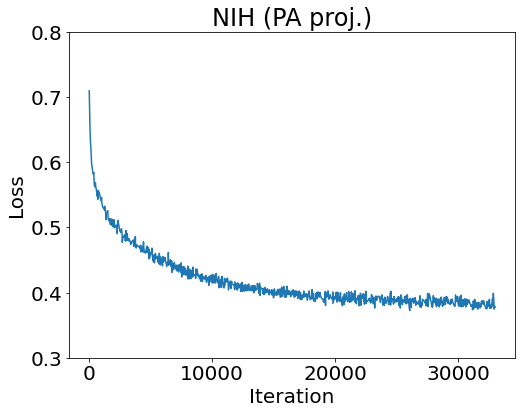}
    \end{subfigure}%
    \hfill
    \begin{subfigure}[]{0.33\textwidth}
        \captionsetup{justification=centering, margin=0cm}
        \centering
        \includegraphics[width=1\textwidth]{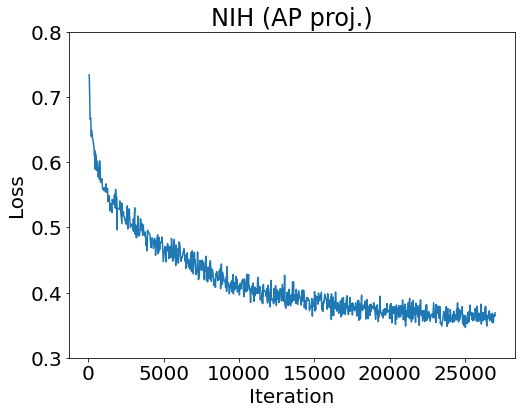}
    \end{subfigure}%
    \caption{\green{Training performance curves of the proposed method on the NIH (a subset), NIH (PA proj.), and NIH (AP proj.) datasets.}}
    \label{fig:loss}
\end{figure*}

\section{Conclusions}

In this manuscript, we evaluated a range of state-of-the-art image anomaly detection methods, the performance of which we found to be sub-optimal in the challenging medical problems.
We proposed a new method that uses an autoencoder to understand the representation of the normal data, with optimization being performed with regard to perceptual loss in the regime of progressive growing training.
To overcome the problem of setting up the model on the new data, we propose to use a small set of anomalous examples of a limited variation \yellow{simply for selecting} the model’s hyperparameters. We believe that this realization reflects real-world clinical scenarios, \yellow{allows} consistent structuring of the experiments, and \yellow{enables} the generation of reproducible results in the future. 
The proposed approach achieved 0.934 ROC AUC in the detection of metastases and 0.926 in the detection of abnormal chest X-rays. Our \yellow{work, thus, establishes} a new strong baseline for anomaly detection in medical imaging.
\\

\appendices

\section{}

Table~\ref{table:naturla_by_classes} \green{reports additional details on average ROC AUC values with standard deviation calculated over 3 runs for every experimental configuration on the CIFAR10 and the SVHN datasets. We considered ten different configurations for each dataset, where one label is designated as normal while the others are considered abnormal.} 

\green{We also show ROC curves for the considered methods on the NIH (a subset), NIH (PA proj.), and NIH (AP proj.) datasets in} Figure~\ref{fig:roc_curves}
and the training performance graph (loss \textit{vs.} iteration) of the proposed method in Figure~\ref{fig:loss}.


{\small
\bibliographystyle{IEEEtran}\
\bibliography{egbib}
}

\begin{IEEEbiography}[{\includegraphics[width=1in,height=1.25in,clip,keepaspectratio]{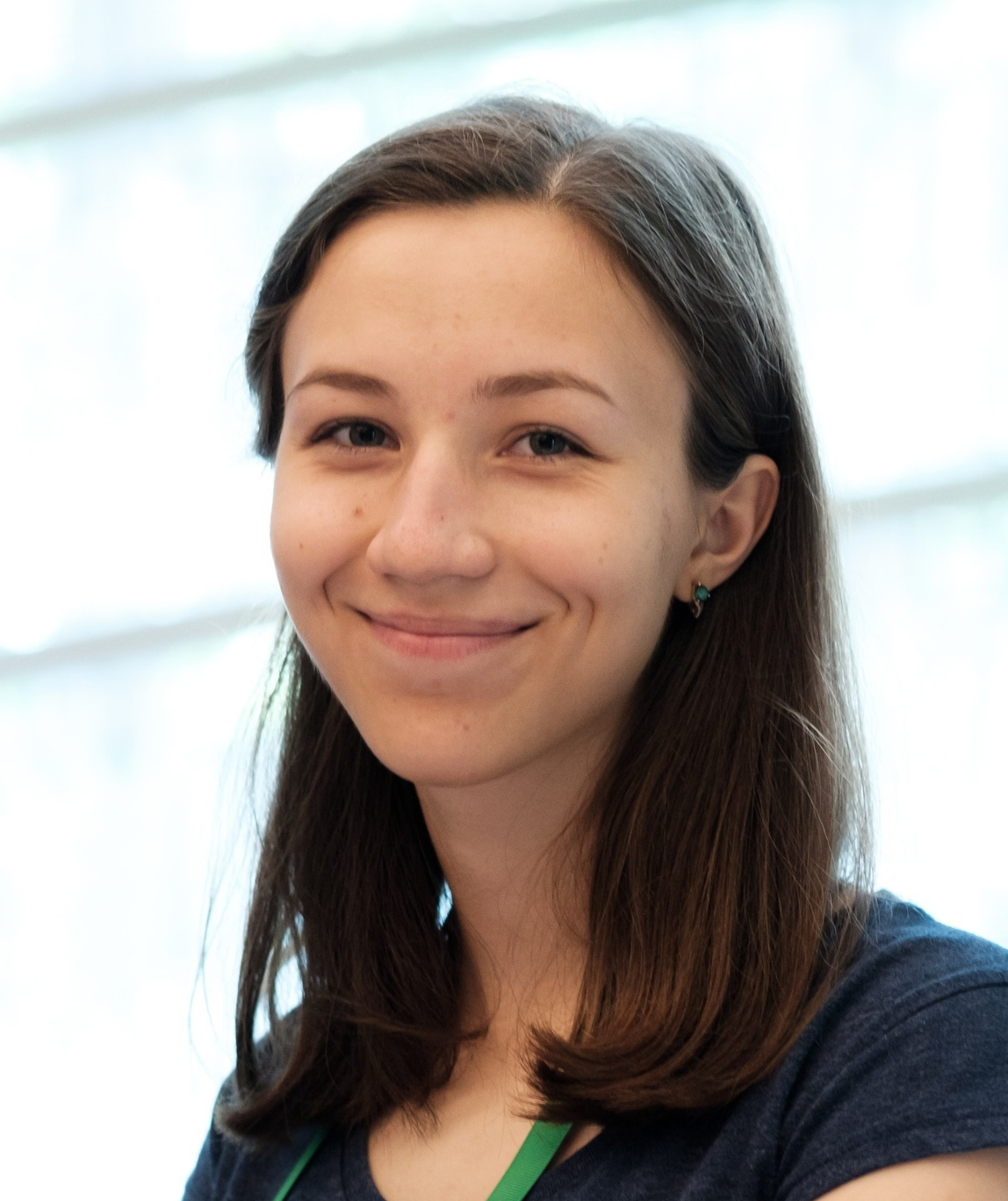}}]{Nina Shvetsova}
received the B.S. and M.S. degree in mathematics and computer science from the Lomonosov Moscow State University, Moscow, Russia in 2019. She is currently pursuing a Ph.D. degree in engineering at the Goethe University Frankfurt, Frankfurt, Germany. 

From 2017 to 2021, she was a Junior Scientist with Philips Research, Moscow, Russia. Her research interests include deep learning, computer vision, and medical image analysis. Since 2021 she is focusing on automated action analysis in video.
\end{IEEEbiography}

\begin{IEEEbiography}[{\includegraphics[width=1in,height=1.25in,clip,keepaspectratio]{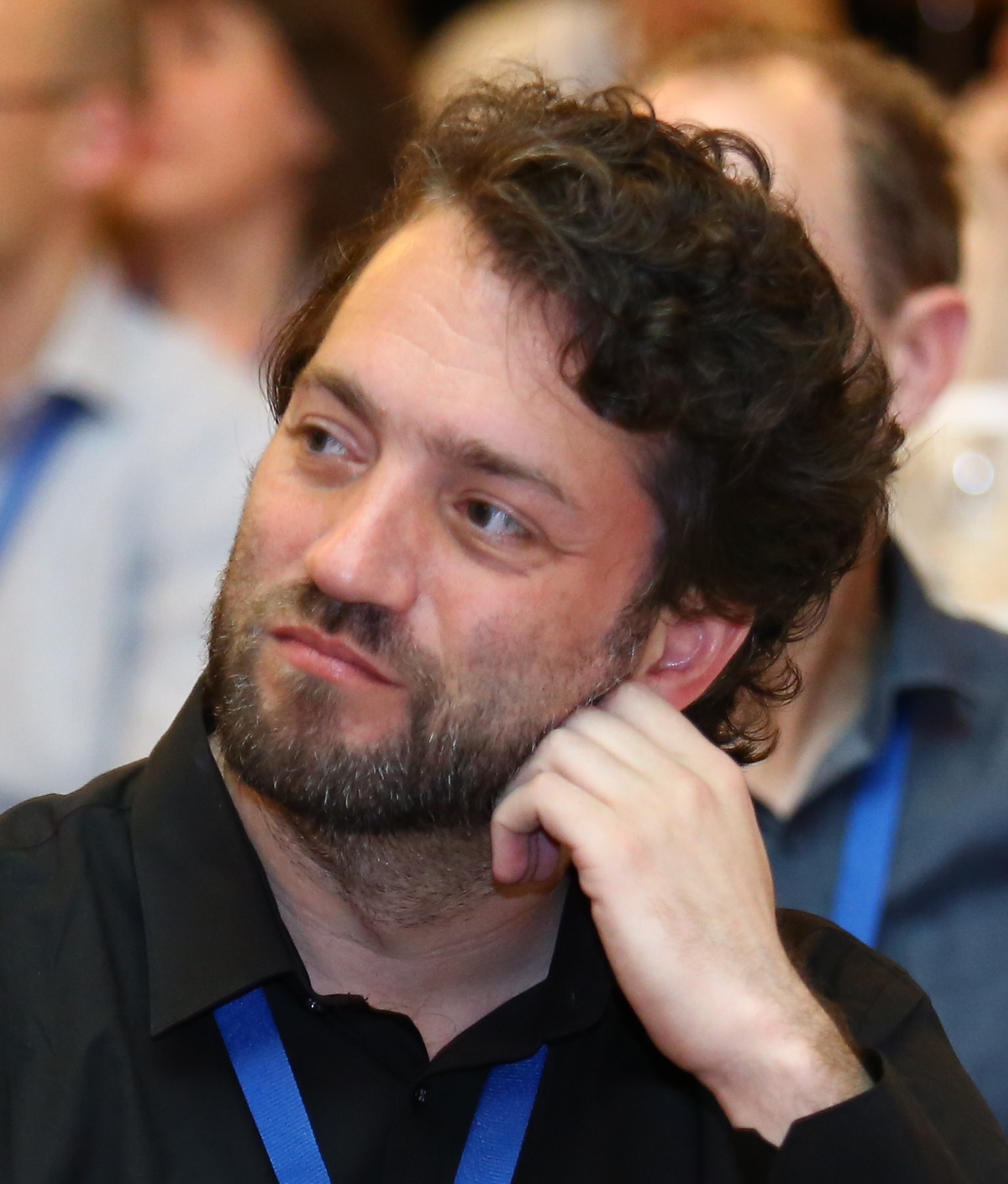}}]{Bart Bakker}
received his M.S degree in computational physics at the Radboud University in Nijmegen, The Netherlands, in 1998. In 2003 he obtained his Ph.D. degree in physics at the same university, based on his research in the area of bayesian statistics and machine learning. 

He has worked for Philips Research since 2003, the first two years as a Marie Curie PostDoc in Aachen, Germany, and since as a Senior Scientist in Eindhoven, The Netherlands. His research interests range from computational modeling of biological systems and machine learning to speech recognition, and since 2016 focus on deep learning and medical image analysis
\end{IEEEbiography}

\begin{IEEEbiography}[{\includegraphics[width=1in,height=1.25in,clip,keepaspectratio]{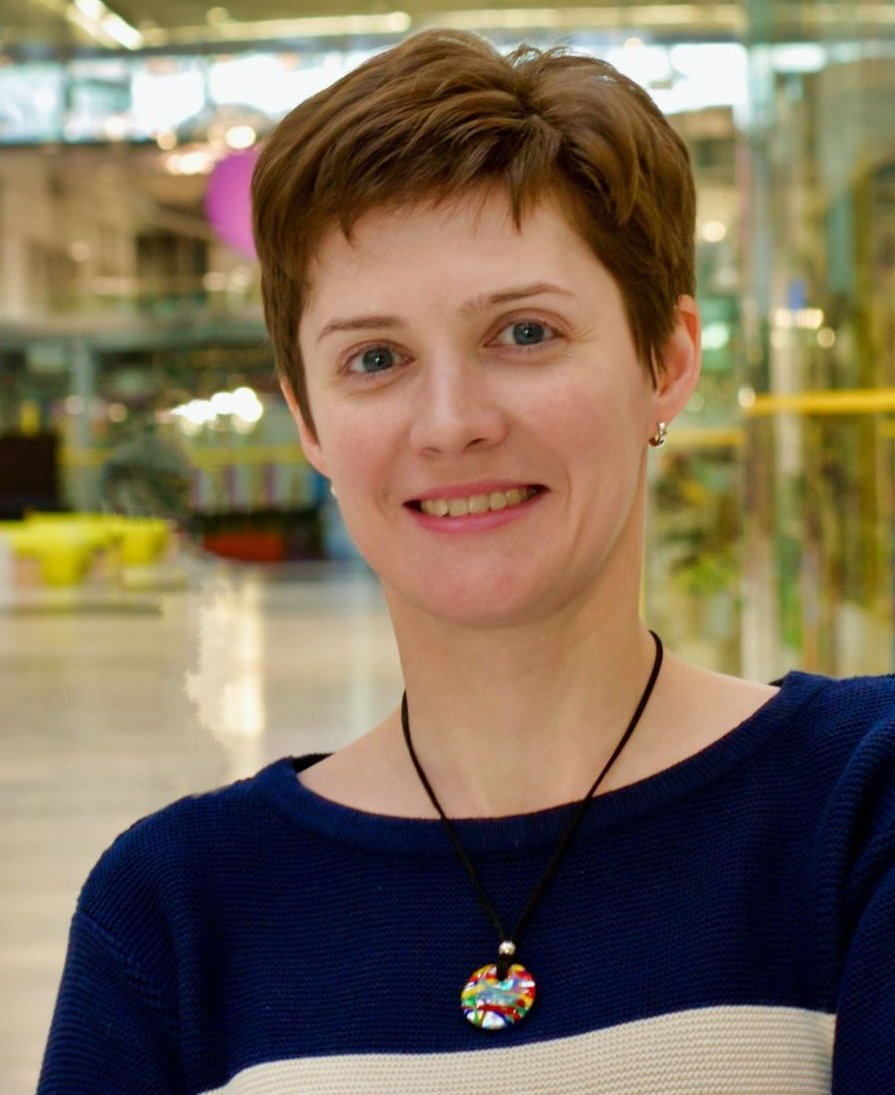}}]{Irina Fedulova}
received the M.S. and Ph.D. degrees in computer science and numerical modeling from Lomonosov Moscow State University, Moscow, Russia in 2004 and 2007 respectively. 

From 2006 till 2017 she worked at IBM, Moscow, Russia as a Staff Software Engineer and Data Scientist, then High-Performance Computing and Data Science Department Manager. Starting from 2017 she joined Philips Research to lead the newly created AI~\&~Data Science research lab in Moscow, Russia. She is passionate about extracting valuable information from data to improve healthcare.
\end{IEEEbiography}

\begin{IEEEbiography}[{\includegraphics[width=1in,height=1.25in,clip,keepaspectratio]{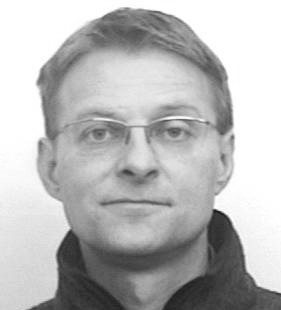}}]{Heinrich Schulz}
obtained a degree in physics from the University of Hamburg in 1995. 

In 1996, he joined Philips Semiconductors starting as a System Architect in semiconductor development. In different positions he moved into digital signal processing and joined Philips Research in 2003. He worked in therapy planning and medical image processing activities. He took over project leader responsibility for research and productization of different image processing projects, in image segmentation, therapy planning and latest in the field of artificial intelligence.
\end{IEEEbiography}

\begin{IEEEbiography}[{\includegraphics[width=1in,height=1.25in,clip,keepaspectratio]{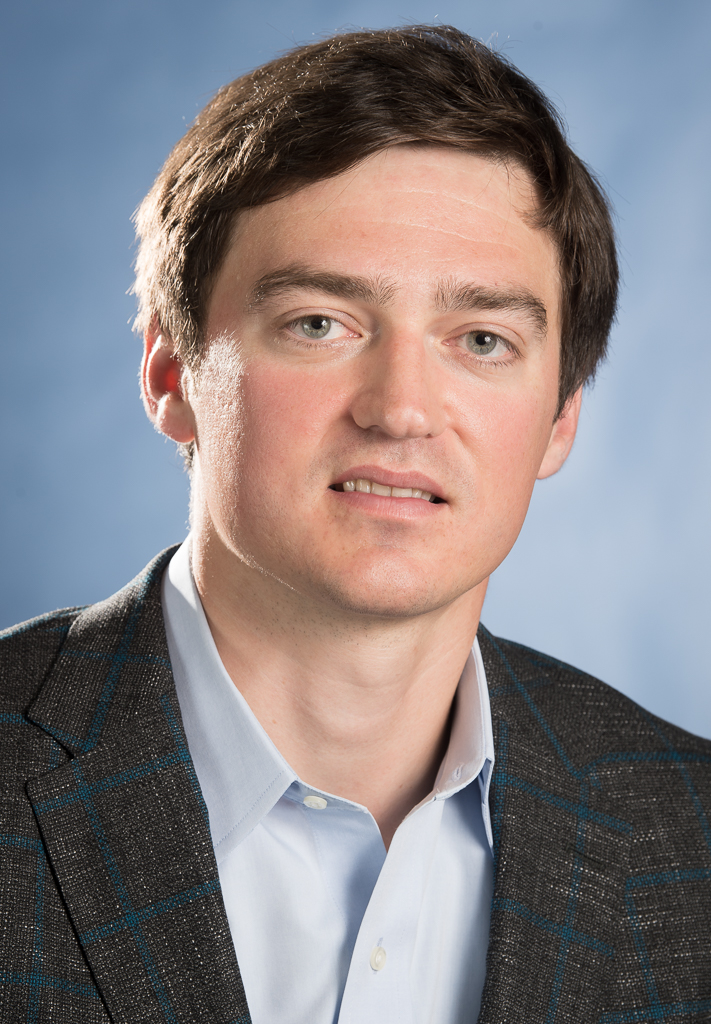}}]{Dmitry V. Dylov}
 received the M.Sc. degree in applied physics and mathematics from Moscow Institute of Physics and Technology, Moscow, Russia, in 2006, and the Ph.D. degree in Electrical Engineering from Princeton University, Princeton, NJ, USA, in 2010. 
 
 He is an Associate Professor and the Head of Computational Imaging Group at Skoltech, Moscow, Russia. His group specializes on computational imaging, computer/medical vision, and fundamental aspects of image formation.\end{IEEEbiography}


\EOD

\end{document}